%% file: main.tex
\documentclass[10pt,twocolumn,letterpaper]{article}

\usepackage[pagenumbers]{cvpr}      
\usepackage{multirow}
\usepackage{makecell}
\usepackage{tabularx}

\usepackage{svg}

\definecolor{cvprblue}{rgb}{0.21,0.49,0.74}
\usepackage[pagebackref,breaklinks,colorlinks,citecolor=cvprblue]{hyperref}

\def\confName{CVPR}
\def\confYear{2024}

\title{ARBiBench: Benchmarking Adversarial Robustness of Binarized Neural Networks}
\author{Peng Zhao$^{1}$, \quad Jiehua Zhang$^{2}$, \quad 
Bowen Peng$^{1}$, \quad Longguang Wang$^{1}$, \\
YingMei Wei$^{1}$, \quad Yu Liu$^{1}$, 
\quad Li Liu$^{1}$\\
\textsuperscript{1} National University of Defense Technology\\
\textsuperscript{2} University of Oulu\\
{\tt\small zhaopeng@nudt.edu.cn, dreamliu2010@gmail.com}
}

\begin{document}
\maketitle
\input{sec/0_abstract}    
\input{sec/1_intro}

\input{sec/2_related_work}
\input{sec/3_evaluation}

\input{sec/4_discussion}

\bibliographystyle{ieeenat_fullname}
\bibliography{main}
\clearpage
\input{sec/6_appendix}



\end{document}

%% file: sec/0_abstract.tex
\begin{abstract}
Network binarization exhibits great potential for deployment on resource-constrained devices due to its low computational cost. Despite the critical importance, the security of binarized neural networks (BNNs) is rarely investigated. In this paper, we present \textbf{ARBiBench}, a comprehensive benchmark to evaluate the robustness of BNNs against adversarial perturbations on CIFAR-10 and ImageNet. We first evaluate the robustness of seven influential BNNs on various white-box and black-box attacks. The results reveal that 1) The adversarial robustness of BNNs exhibits a completely opposite performance on the two datasets under white-box attacks. 2) BNNs consistently exhibit better adversarial robustness under black-box attacks. 3) Different BNNs exhibit certain similarities in their robustness performance. Then, we conduct experiments to analyze the adversarial robustness of BNNs based on these insights. Our research contributes to inspiring future research on enhancing the robustness of BNNs and advancing their application in real-world scenarios.

\end{abstract}

%% file: sec/1_intro.tex
\section{Introduction}
\label{sec:intro}
Deep neural networks (DNNs) have demonstrated remarkable performance in numerous computer vision tasks, including image classification \cite{intro1,intro2,intro3, MPFGVC}, object detection \cite{intro4,intro5,intro6,intro7}, semantic segmentation \cite{intro8,intro9}, \textit{etc}. Nonetheless, DNNs produce promising results at the cost of large model sizes and high computational complexity, which limit their practical deployment on resource-constrained devices. To address this challenge, researchers have developed various model compression techniques, including network quantization \cite{intro10,intro11,intro12, su2022svnet}, network pruning \cite{intro13,intro14,intro15}, knowledge distillation \cite{intro16,intro17,intro18}, and low-rank matrix factorization \cite{intro19,intro20}. Among these techniques, Binary Neural Networks (BNNs) can be regarded as an extreme case of quantization \cite{intro21,intro22}, which achieve significant model acceleration by quantizing weights and activations to +1 and -1.


\begin{table*}[ht]
\centering

\renewcommand{\arraystretch}{1.32} 
\resizebox{\linewidth}{!}{
\begin{tabular}{lcccl}
\toprule

Paper   & White-box Attacks & Black-box Attacks & Target Model & \multicolumn{1}{c}{Main Content}     \\ \hline

ICLR2018 \cite{attack_BNN1} & \makecell[c]{5}                         & 1  &  BNN \cite{BNN} &   First to evaluate and interpret the adversarial robustness of BNNs.                                                                 \\
ICLR2019 \cite{attack_BNN2}   & 3           &0 & Quantized Models &  Bridge model quantization and adversarial defense with Defensive Quantization.                                                                                 \\
arXiv2021 \cite{attack_BNN5} & \makecell[c]{4}                         &2 & Efficient CNN & Study the robustness of distilled,
pruned and binarized neural networks.                                                                 \\
AAAI2022 \cite{attack_BNN4} &2                        & 1 & Fully BNNs  &  Investigate the issue of gradient vanishing in fully BNNs under attacks.                                                                                  \\

arXiv2023 \cite{DBLP:journals/corr/abs-2308-02350} & 3                          & 0 & \makecell[l]{Three 
Quantization methods \cite{DoReFa,choi2018pact,esser2019learned}}     & Comprehensively evaluate the robustness of quantized models (2, 4, 6, 8 bits).                                                                           \\ \hline
ARBiBench (ours)  & \makecell[c]{4}              & \makecell[c]{6}  &Seven SOTA binarization methods & Comprehensively evaluate the robustness of 7 binarization methods.                                                                                \\ \hline
\end{tabular}
}
\caption{Compared with previous works on adversarial robustness for quantized neural networks. We present the number of white-box and black-box attack methods, the target model investigated, and the main content for each work. }
\label{Comparsion}
\end{table*}
While benefiting from efficient computation and reduced storage requirements, the trustworthiness of BNNs needs to be further validated when exposed to various perturbations in real-world scenarios, such as adversarial attacks and noises. In this paper, we study the robustness of BNNs against adversarial attacks. Previous works suggest \cite{firstadv,easyattack2} that 32-bit floating-point networks (FP32) are fragile to adversarial samples. Compared with FP32, researchers explore the robustness of quantized neural networks and suggest that they exhibit better adversarial robustness, and increasing the quantization bit-width generally leads to a decrease in adversarial robustness \cite{DBLP:journals/corr/abs-2308-02350}. These studies primarily analyzed the robustness of low-bit quantized neural networks from 2-bit to 8-bit and did not consider BNNs. While prior researches \cite{attack_BNN1,attack_BNN2,attack_BNN3,attack_BNN4,attack_BNN5} have explored the adversarial robustness of BNNs, a comprehensive evaluation like \cite{croce2020robustbench,DBLP:journals/corr/abs-2308-02350} has not been thoroughly conducted on BNNs, leading to a gap in understanding their vulnerability and potential benefits. 



To bridge this gap, we present \textbf{ARBiBench}, a comprehensive benchmark designed to evaluate the adversarial robustness of BNNs. \textbf{ARBiBench} comprehensively evaluates the adversarial robustness of seven representative binarization methods on CIFAR-10 and ImageNet. To provide a thorough investigation of the adversarial robustness of BNNs, We evaluate these models using ten distinct adversarial attack methods, consisting of four white-box attacks and six black-box attacks under $\ell_2$ and $\ell_\infty$ norm. Tab. \ref{Comparsion} highlights the distinction between \textbf{ARBiBench} and existing approaches in analyzing the robustness of quantized neural networks. Our empirical results illustrate that different BNNs exhibit certain similarities in the adversarial robustness performance. Under white-box attacks, the adversarial robustness of BNNs exhibits a completely opposite performance on small-scale and large-scale datasets compared with FP32. For black-box attacks, BNNs exhibit greater adversarial robustness than FP32. Furthermore, we conduct experiments to analyze the observed phenomena, facilitating a deeper understanding of the adversarial robustness of BNNs. Our contributions can be summarized as follows:

\begin{itemize}
    \item To the best of our knowledge, ARBiBench is the first work that systematically evaluates the adversarial robustness of BNNs. ARBiBench evaluates the adversarial robustness of seven representative BNNs on both small- and large-scale datasets using ten attack methods, including white-box attacks and various types of black-box attacks.
    \item Through extensive experimentation, ARBiBench challenges the previous singular perspective on the robustness of BNNs and reveals crucial insights into the robustness of BNNs under both white-box and black-box attacks. 
    
    \item ARBiBench provides a comprehensive benchmark for evaluating the robustness of BNNs, promoting further research on the security in the practical deployment of BNNs.
    
\end{itemize}

%% file: sec/2_related_work.tex
\section{Related Works}
In this section, we provide a comprehensive review of the recent works on BNNs, adversarial attacks, and the advancements in their integration.
\subsection{Binary Neural Networks}
BNNs \cite{DBLP:conf/nips/HubaraCSEB16} is an extreme case of model quantization that constrains weights and activations to -1 or +1 using the sign function. This unique approach allows BNNs to leverage bitwise operations, such as the xnor and popcount operations, reducing the computational complexity significantly. The main pipeline of forward propagation can be presented as follows: given weights $\boldsymbol{w} \in \mathbb{R}^{c_{\text {in }} \times c_{\text {out }} \times k \times k}$ and activations $\boldsymbol{a} \in \mathbb{R}^{c_{\text {in }} \times w \times h}$, where $c_{\mathrm{in}}, k, c_{\mathrm{out}}, w$, and $h$ denote the input channel, kernel size, output channel, input width, and input height. The computation can be expressed as
\begin{equation}
\boldsymbol{o}= (\operatorname{sign}(\boldsymbol{a}) \circledast \operatorname{sign}(\boldsymbol{w})) \odot \alpha,
\end{equation}
where $\boldsymbol{o}$ denotes the outputs, $\circledast$ denotes the bit-wise operations including xnor and popcount, $\odot$ denotes the element-wise multiplication, and $\alpha$ denotes optional scaling factor. \cite{XNOR} demonstrated that BNNs can bring 32× storage compression ratio and 58× practical computational reduction on CPU. As the sign function is non-differentiable, BNNs employ surrogate functions (\textit{e.g.}, STE \cite{DBLP:STE}) during the backpropagation for gradient computation.


There are two major challenges \cite{suzhuo} in BNNs: gradient mismatch and large quantization errors. Recent works primarily aim to address these two issues. To tackle the first challenge, various differentiable functions have been proposed to estimate the gradient of the sign function \cite{Bi-Real, ir-net, frequency}. Besides, auxiliary modules were proposed to facilitate the optimization of binary weights \cite{auxiliary1, auxiliary2}, and specific regularization techniques were employed in activations to guide gradient computations \cite{regularization1, regularization2}.
Regarding the second challenge, some studies introduced scaling factors to reduce the quantization error of weights and activations. The scale factors can be pre-computed \cite{XNOR, DoReFa} or learnable \cite{XNOR++, learned1, learned2}, and significantly improve the representation of BNNs. Furthermore, novel optimization and training methods are also helpful to mitigate quantization errors \cite{ReCU,optimization1,optimization2}.

\subsection{Adversarial Attacks}\label{adversarial attacks}
Given the fact that DNNs are severely vulnerable to adversarial examples \cite{easyattack2, firstadv}, where clean inputs are added with imperceptible yet deliberately designed adversarial perturbations. A huge amount of work has been dedicated to evaluating \cite{tang2021robustart,croce2020robustbench}, finding \cite{pgd,cw}, and defending \cite{feinman2017detecting,tramer2017ensemble,raghunathan2018certified} adversarial examples. This paper focuses on evaluating the adversarial vulnerability of BNNs in widely investigated settings and threat models. In specific, we limit our scope to widely investigated setting, \textit{i.e.}, making models output wrong predictions via adding $\ell_\infty$- and $\ell_2$-restricted perturbations other than specifically targeted ones and other distance metrics \cite{engstrom2019exploring,laidlaw2019functional,wong2019wasserstein}. We follow \cite{Dong1} in categorizing existing attacks based on the adversaries' knowledge, and the main studied attacks are white-box attacks, score-based and decision-based attacks, and transfer attacks.


The goal of the adversary is generally formulated as
\begin{equation}
    \centering
    \underset{\delta:\|\delta\|_{p}\leq\epsilon}{\operatorname{maximize}} \,\, \mathcal{L}(C(x+\delta),y),
    \label{atkobj}
\end{equation}
where $C: \mathcal{X} \to \mathcal{Y}$ represents the classifier that maps input $\boldsymbol{x}\in\mathcal{X}$ to label space $y\in\mathcal{Y}$.
We choose the cross-entropy loss function as $\mathcal{L}(\cdot)$ for the classification task. The adversarial perturbation $\delta$ falls into the $\epsilon$-bounded $\ell_{p}$ ball centered at the clean input $x$ to ensure imperceptibility. 
\textbf{White-box attacks} utilize full information of the target model for crafting perturbations, including architecture, parameters, and gradients. In this scenario, adversaries are able to exploit the gradients to find feasible solutions for objective \eqref{atkobj}. For instance, Fast Gradient Sign Method (FGSM) \cite{firstadv} makes a $\epsilon$-long step along the ascending direction of gradients. Projected Gradient Descent (PGD) \cite{pgd} opts to generate perturbation from random initialization in an iterative manner with a small step size. The aforementioned methods are known as effective yet highly efficient and are readily applied for both $\ell_{\infty}$ and $\ell_{2}$ restrictions. Another branch of work in our focus manages to pursue both classification error and minimum perturbation magnitude, such as DeepFool \cite{deepfool} and C\&W \cite{cw}.

It turns to the black-box threat model when access to gradients of $C$ is prohibited while the output $C(x+\delta)$ can be queried, which aligns well with the MLaaS operating scenario. Then the attack goal can be solved by gradient-free black-box optimization approaches. \textbf{Score-based attacks} exploits the soft output to iteratively update perturbations. SPSA \cite{spsa} estimates the gradient of the target model by applying random perturbations. $\mathcal{N}$ATTACK \cite{nattack} focuses on learning a Gaussian distribution centered around the input with the assumption that any sample drawn from this distribution is likely to be a good candidate for an adversarial example. Square \cite{square} employs a stochastic search strategy. It selects local square updates for perturbations, which enhances the query efficiency during the attack process. \textbf{Decision-based attacks} focus on crafting perturbations solely depending on the hard label of prediction. Boundary \cite{boundary} samples from a normal distribution to find an ideal perturbation around the decision boundary. Evolutionary \cite{evol} enhances query efficiency for generating adversarial examples by employing an evolution strategy.
\textbf{Transfer-based attacks} aim to boost the transferability of adversarial examples towards successfully attacking black-box models from a local surrogate model. The surrogate and target models are different such as in weights, architecture, and training strategy. Many methods were proposed to alleviate the overfitting of the surrogate model, including introducing label-preservation transformations like resizing \cite{transfer3}, translation \cite{transfer4}, and scaling \cite{transfer5}, \textit{etc}, and advanced gradient calculations such as momentum \cite{transfer1} and Nesterov acceleration \cite{transfer5}. We select the Scale-Invariant Nesterov Iterative Fast Gradient Sign Method (SI-NI-FGSM) \cite{transfer5} for transfer attack evaluation as it typically modifies the baseline method on both sides and performs well in previous studies.

\subsection{Robustness of BNNs}

A large number of works have provided a comprehensive study on the robustness of FP32 and quantized models \cite{tang2021robustart,croce2020robustbench,DBLP:journals/corr/abs-2308-02350,vora2023benchmarking}. However, the robustness of BNNs has not been fully explored. Galloway \textit{et al.} \cite{attack_BNN1} firstly evaluated and interpreted the robustness of BNNs to adversarial attacks. They introduced that BNNs enhance robustness against attacks by introducing stochasticity. However, this approach has been shown to exhibit fake robustness \cite{false_gradient}. Lin \textit{et al.} \cite{attack_BNN2} observed that quantized models are more susceptible to adversarial attacks compared with FP32, particularly when the quantization bit width exceeds 3 bits. Nevertheless, Vemparala \textit{et al.} \cite{attack_BNN5} argued that BNNs exhibit resilient behavior compared to their baselines and other compressed variants. Gupta \textit{et al.} \cite{attack_BNN4} revealed that quantized models suffer from gradient vanishing issues and show a fake sense of robustness. In this paper, we aim to comprehensively evaluate the robustness of BNNs against multiple adversarial attacks for several binarization methods on small- and large-scale datasets.



%% file: sec/3_evaluation.tex
\section{Evaluation Metrics and Settings}
In this section, we will introduce some of the metrics used in the experiments, as well as some fundamental experimental settings.
\subsection{Evaluation Settings}

\textbf{Binarization Methods}: We consider ResNet18 as our base model. For binarization methods, we focus on seven popular techniques, including BNN \cite{BNN}, XNOR-Net \cite{XNOR}, DoReFa \cite{DoReFa}, Bi-Real Net \cite{Bi-Real}, XNOR-Net++ \cite{XNOR++}, ReActNet \cite{ReAct} and ReCU \cite{ReCU}. The details of these methods are presented in Appendix A. We follow the BiBench \cite{Bibench} to implement these methods. 


\textbf{Adversarial Attacks}: In ARBiBench, we employ 10 attack methods, including four white-box attacks (FGSM \cite{firstadv}, PGD \cite{pgd}, DeepFool \cite{pgd}, and  C\&W \cite{cw}), three score-based attacks (SPSA \cite{spsa}, $\mathcal{N}$ATTACK \cite{nattack}, and Square \cite{square}), two decision-based attacks (Boundary \cite{boundary} and Evolutionary \cite{evol}), and one transfer-based attacks (SI-NI-FGSM \cite{transfer5}). For each attack, we set different perturbation budgets to evaluate the robustness of BNNs.


\textbf{Training Setting}: We train ResNet18 with FP32 and seven binarization methods on CIFAR-10 and ImageNet dataset. For CIFAR-10, we employ the following training parameters: 80 epochs, Adam optimizer, the multistep learning rate strategy with steps at 50 and 75, and the initial learning rate of 0.001.  For ImageNet, the training parameters are as follows: 120 epochs, SGD optimizer, the cosine learning rate strategy with an initial learning rate of 0.1, the warm-up period of 10 epochs, and the weight decay of 2e-5.


\textbf{Evaluation Setting}: 
Following the previous method \cite{Dong1}, we adopt \textit{accuracy vs. perturbation budget} curve to clearly and thoroughly show the robustness and resistance of the model against the attack. The perturbation budget denotes varying levels of perturbation. The accuracy is normalized adversarial accuracy under the perturbation budget. The detailed information can be found in Sec. \ref{sub: metric}. For evaluated datasets, we use CIFAR-10 and ImageNet to evaluate adversarial robustness. For CIFAR-10, we use the entire test set for white-box attacks and 500 randomly selected images from the test set for black-box attacks. For ImageNet, 1000 random images from the test set are used across all attack methods following \cite{Dong1}. For attack techniques without inherent perturbation constraints, like CW and DeepFool, we impose constraints on the generated adversarial examples.



\subsection{Evaluated Metrics}
\label{sub: metric}
Given a classifier $C$ defined in Sec. \ref{adversarial attacks}. The accuracy of the classifier is defined as
\begin{equation}\label{eq:accuracy}
ACC = \frac{1}{N} \sum_{i=1}^N \mathbf{1}\left(C\left(x_i\right) = y_i\right),
\end{equation}
where $\left\{\boldsymbol{x}_i, y_i\right\}_{i=1}^N$ is the test set, $\mathbf{1}(\cdot)$ is the indicator function. For specific adversarial attacks, the accuracy of the classifier against the attack can be calculated with the following expression:
\begin{equation}\label{eq:adv_accuracy}
ACC^*=\frac{1}{N} \sum_{i=1}^N \mathbf{1}\left(C\left(\mathcal{A}_{\epsilon, p}\left(x_i\right)\right) = y_i\right),
\end{equation}
where $\mathcal{A}_{\epsilon, p}$ represents the attack method, $\epsilon$ stands for the perturbation budget and $p$ indicates the distance norm. Traditionally, adversarial robustness ($AR$) is measured by its accuracy on adversarial examples ($ACC^*$), with higher $ACC^*$ values indicating greater robustness. However, we aim to evaluate the robustness among models with different clean accuracy. We used the normalized adversarial accuracy to measure the relative performance, which is calculated as follows:
\begin{equation}
ACC_{norm}=\frac{ACC^*}{ACC},
\end{equation} 
Considering a union of different attacks, We employ the mean of $ACC_{norm}$ as a Robustness Score (RS) to assess the adversarial robustness (higher values indicate a more robust model):
\begin{equation}
RS = \sum_{i=1}^N ACC_{\text {norm}}^i,
\end{equation}
where $N$ is the number of attack methods. Note that we utilize accuracy to denote the normalized adversarial accuracy in subsequent sections for convenience.

\section{Evaluation Results}
This section reports our evaluation results, including the results on CIFAR-10 in Sec. \ref{cifar10} and the results on ImageNet in Sec. \ref{imagenet}. The robustness results of defense approaches are reported in Appendix B, and the detail of clean accuracy is reported in Appendix C. We report the results of black-box attacks with $\ell_\infty$ norm constraint, and white-box attacks are subjected to both the $\ell_\infty$ and $\ell_2$ norm constraints.



\subsection{Results on CIFAR-10}\label{cifar10}

\subsubsection{White-box Attacks}
\begin{figure*}[t]
\begin{minipage}{.57\linewidth}
\begin{center}
\includegraphics[width=1.0\linewidth]{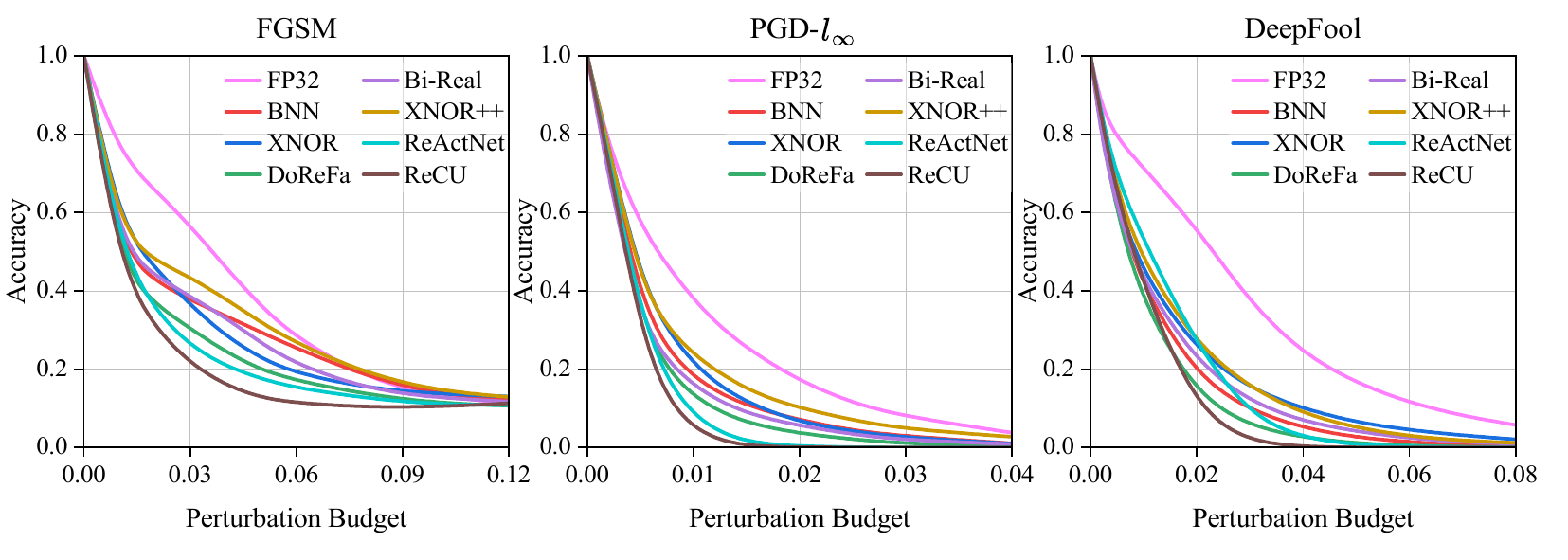}
\end{center}
\vspace{-3ex}
\caption{The \textit{accuracy vs. perturbation budget} curves of the $8$ normally trained models on CIFAR-10 against untargeted white-box attacks under the $\ell_{\infty}$ norm.}
\label{fig:whitelinf-ut-linf-cifar10-acc-pert}
\end{minipage}
\hspace{1ex}
\begin{minipage}{.41\linewidth}
\begin{center}
\includegraphics[width=0.9\linewidth]{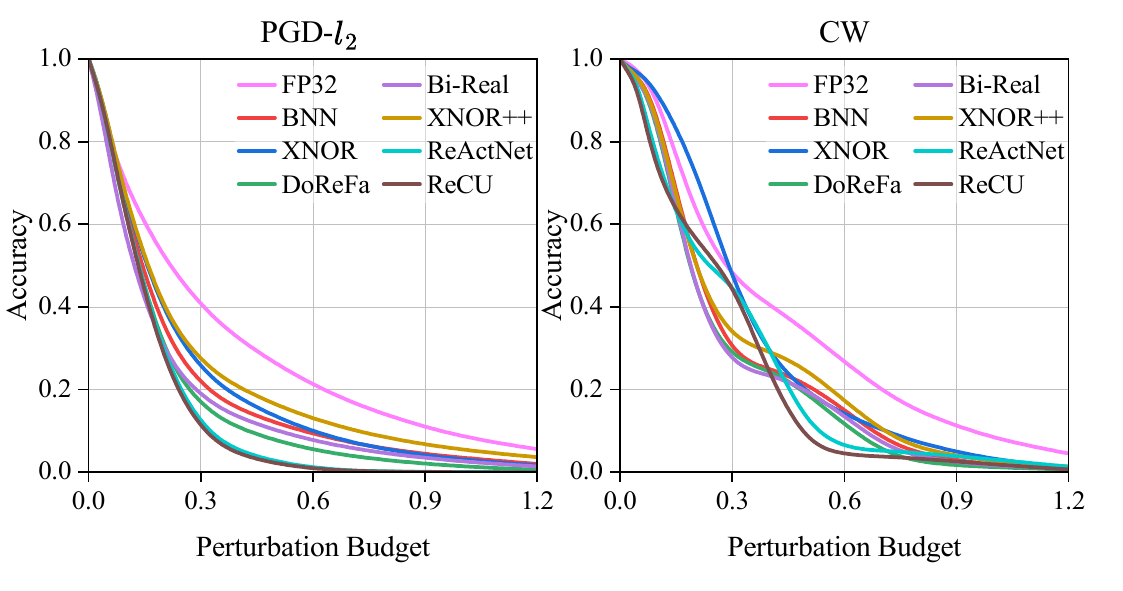}
\end{center}
\vspace{-3ex}
\caption{The \textit{accuracy vs. perturbation budget} curves of the $8$ normally trained models on CIFAR-10 against untargeted white-box attacks under the $\ell_2$ norm.}
\label{fig:whitel2-ut-l2-cifar10-acc-pert}
\end{minipage}
\begin{minipage}{.57\linewidth}
\begin{center}
\includegraphics[width=1.0\linewidth]{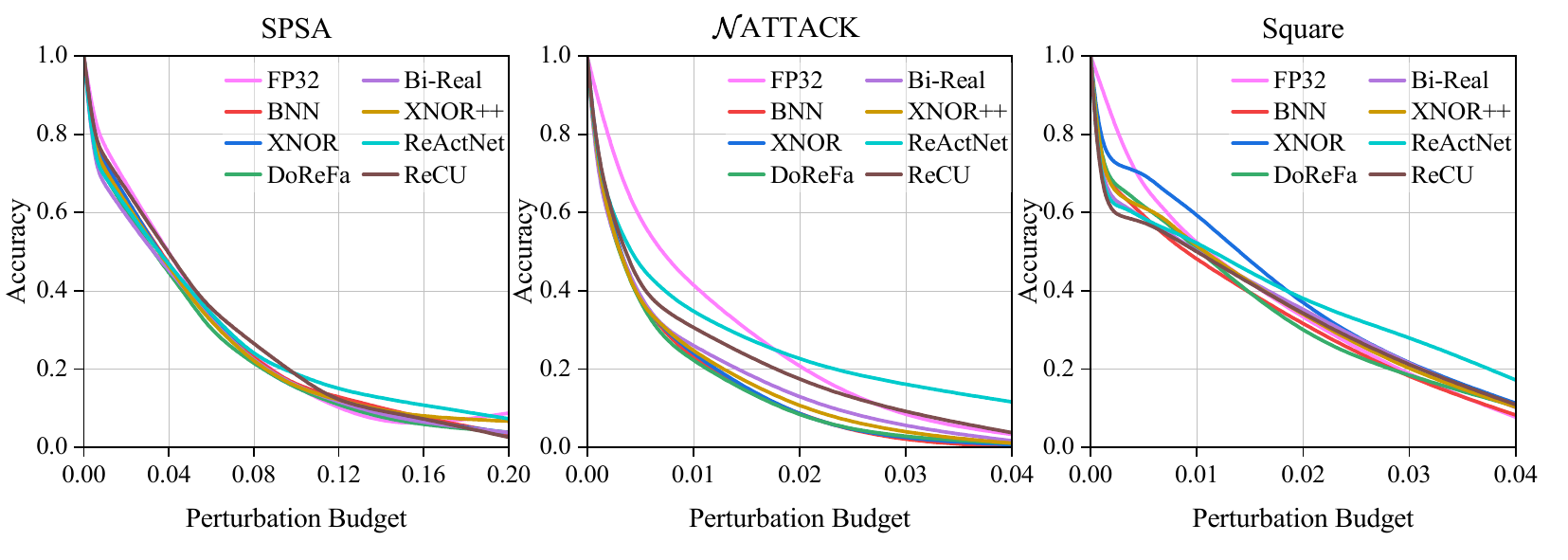}
\end{center}
\vspace{-3ex}
\caption{The \textit{accuracy vs. perturbation budget} curves of the $8$ normally trained models on CIFAR-10 against untargeted score-based attacks under the $\ell_{\infty}$ norm.}
\label{fig:score-ut-linf-cifar10-acc-iter}
\end{minipage}
\hspace{1ex}
\begin{minipage}{.41\linewidth}
\begin{center}
\includegraphics[width=0.9\linewidth]{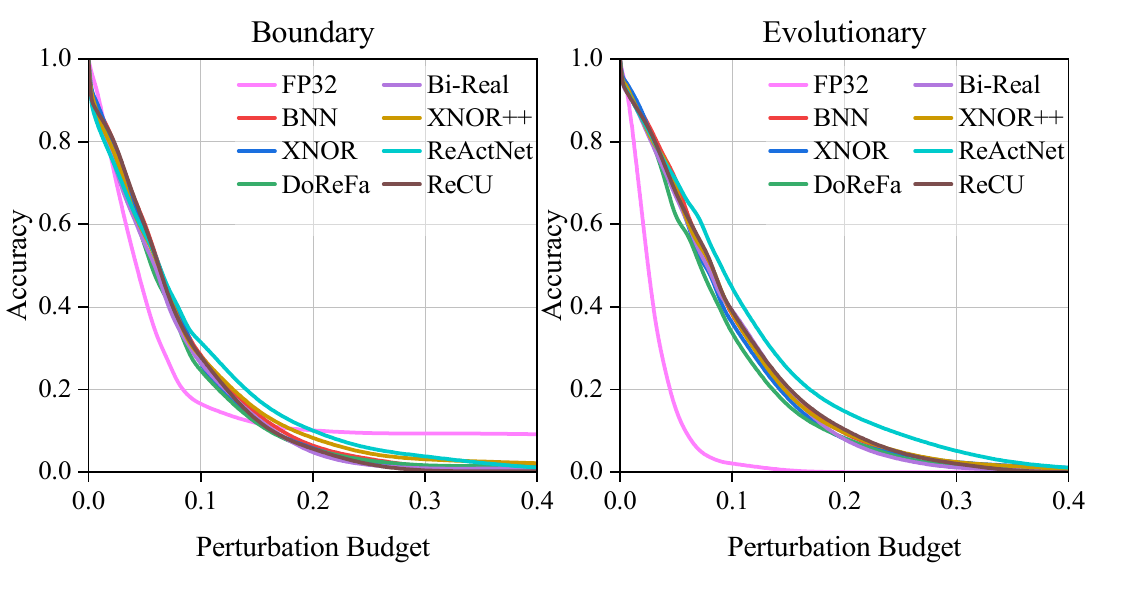}
\end{center}
\vspace{-3ex}
\caption{The \textit{accuracy vs. perturbation budget} curves of the $8$ normally trained models on CIFAR-10 against untargeted decision-based attacks under the $\ell_{\infty}$ norm.}
\label{fig:decision-ut-l2-cifar10-acc-iter}
\end{minipage}
\end{figure*}

We present the \textit{accuracy vs. perturbation budget} curves for evaluating the robustness of 8 models against FGSM, PGD, and DeepFool attacks under the $\ell_\infty$ norm in Fig. \ref{fig:whitelinf-ut-linf-cifar10-acc-pert}, and for PGD and C\&W attacks under the $\ell_2$ norm in Fig. \ref{fig:whitel2-ut-l2-cifar10-acc-pert}, from which the following observations can be drawn. \textbf{First}, as the perturbation budget is increased, the white-box attacks result in accuracy near zero, but FGSM yields about 10\% for all models. We attribute it to the single-step, weak manipulation power of FGSM. And with larger budgets, it converges to a state of random selection (1/10). \textbf{Second}, BNNs exhibit clearly less robustness than the FP32 model in the white-box threat model, and there is only one exception, the XNOR model in C\&W attack curve (right of Fig. \ref{fig:whitel2-ut-l2-cifar10-acc-pert}) at lower $\ell_{2}$ budgets. Tab. \ref{tab:tab1} also demonstrates that under white-box attacks with a fixed budget, the robustness of BNNs is consistently weaker than that of FP32. \textbf{Third}, strong robustness is not guaranteed by better generalization performance, at least, these two metrics do not always align well. We find that the SOTA binarization method ReCU exhibits the worst robustness in almost all white-box adversarial scenarios, while XNOR++ achieves the highest robustness among these five attacks. This raises demand for a special focus on the robustness in designing BNNs.

\begin{table*}[htbp]
  \centering
    \small
    \begin{tabular}{cccccccccc}
    \toprule
    \multicolumn{2}{c}{Attack} & FP32  & BNN   & XNOR  & DoReFa & Bi-Real & XNOR++ & ReActNet & ReCU \\
    \midrule
    \midrule
    \multirow{6}{*}{White} & FGSM  & \textbf{56.68\%} & 37.66\% & 36.38\% & 30.43\% & 38.59\% & 43.56\% & 25.98\% & 14.54\% \\
          & PGD($\ell_\infty$)   & \textbf{7.87\%} & 2.83\% & 2.47\% & 0.96\% & 1.90\% & 4.64\% & 0.00\% & 0.00\% \\
          & DeepFool & \textbf{36.97\%} & 9.45\% & 14.99\% & 5.37\% & 11.63\% & 15.16\% & 8.00\% & 1.37\% \\
          & PGD($\ell_2$) & \textbf{25.17\%} & 11.35\% & 12.64\% & 7.15\% & 9.59\% & 15.62\% & 2.21\% & 1.04\% \\
          & CW    & \textbf{34.11\%} & 21.38\% & 19.14\% & 19.16\% & 19.96\% & 24.90\% & 11.04\% & 6.56\% \\
          & \textbf{Robustness Score} & \textbf{32.16\%} & 16.53\% & 17.13\% & 12.61\% & 16.33\% & 20.77\% & 9.45\% & 6.08\% \\
    \midrule
    \multirow{4}{*}{Score} & SPSA  & \textbf{58.46\%} & 54.59\% & 54.55\% & 51.72\% & 51.49\% & 53.88\% & 53.33\% & 57.55\% \\
          & $\mathcal{N}$ATTACK & 7.10\% & 1.34\% & 2.00\% & 2.36\% & 5.11\% & 3.45\% & \textbf{15.91\%} & 8.75\% \\
          & Square & 18.20\% & 17.48\% & 20.88\% & 18.06\% & 20.89\% & 19.44\% & \textbf{28.44\%} & 20.61\% \\
          & \textbf{Robustness Score} & 27.92\% & 24.47\% & 25.81\% & 24.05\% & 25.83\% & 25.59\% & \textbf{32.56\%} & 28.97\% \\
    \midrule
    \multirow{3}{*}{Decision} & Boundary & 63.67\% & \textbf{74.50\%} & 71.40\% & 69.96\% & 68.09\% & 72.20\% & 69.89\% & 74.84\% \\
          & Evolutionary & 37.79\% & \textbf{81.66\%} & 79.38\% & 79.40\% & 78.51\% & 79.10\% & 80.43\% & 80.96\% \\
          & \textbf{Robustness Score} & 50.73\% & \textbf{78.08\%} & 75.39\% & 74.68\% & 73.30\% & 75.65\% & 75.16\% & 77.90\% \\
    \bottomrule
    \end{tabular}%
    \caption{The point-wise results of 8 normally trained models on CIFAR-10 against untargeted attacks under $\ell_\infty$ norm for fixed $\epsilon=0.03$ and $\ell_2$ norm for fixed $\epsilon=0.5$.}
  \label{tab:tab1}%
\end{table*}%

\subsubsection{Black-box Attacks}
\textbf{Score-based Attacks}: Fig. \ref{fig:score-ut-linf-cifar10-acc-iter} shows the \textit{accuracy vs. perturbation budget} curves of models against the score-based attacks. We fix the query number as 10,000 for both SPSA and $\mathcal{N}$ATTACK and 5,000 for Square. The SPSA shows the weakest yet consistent attack effects for all models. In a comprehensive appraisal, the disparities in robustness across the various networks are relatively modest.

\textbf{Decision-based Attacks}: Fig. \ref{fig:decision-ut-l2-cifar10-acc-iter} illustrate the \textit{accuracy vs. perturbation budget} curves of models against the decision-based attacks. We fix the query number as 10,000 for both Boundary and Evolutionary in evaluation. The decision-based attacks require a larger perturbation budget to mount effective attacks due to the limited access to models. In this scenario, the BNNs are somehow shown to be more robust than FP32, especially when facing the evolutionary attack. Note that, in terms of Boundary attack, we attach importance to the results where $\epsilon<0.15$ since budgets larger than $38/255$ are less meaningful for defining an adversarial perturbation.

\begin{figure}[ht]
  \centering
  \includegraphics[width=\linewidth]{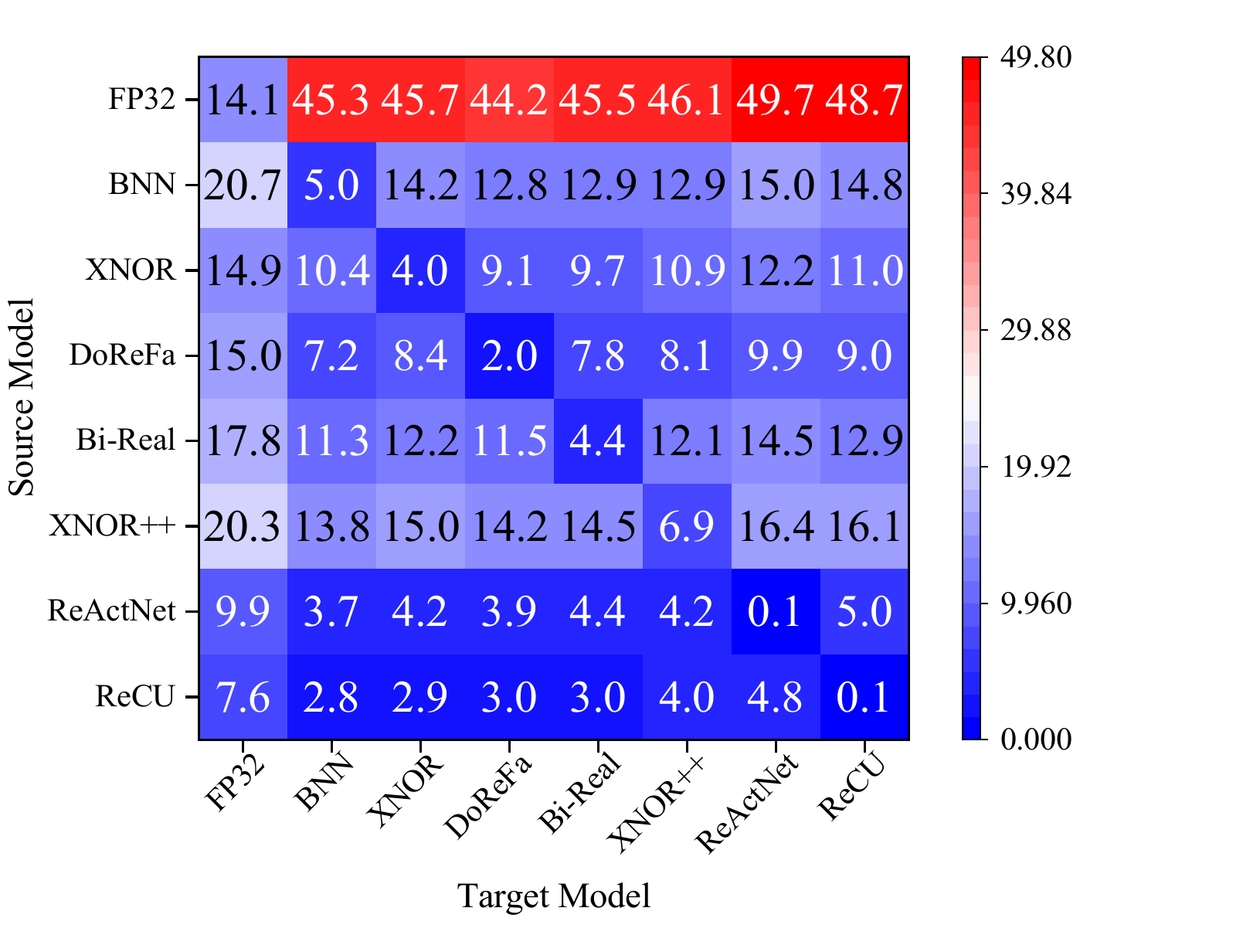}
  \caption{The heatmap of the 8 normally trained models on CIFAR-10 against SI-NI-FGSM attacks under the $\ell_{\infty}$ norm.}
  \label{fig:tranfer_normal}
\end{figure}

\textbf{Transfer-based Attacks}: We use SI-NI-FGSM (with $\epsilon=0.03$ for $\ell_{\infty}$ norm) to generate adversarial examples for each model and perform transfer attack with results shown in Fig. \ref{fig:tranfer_normal}. It is intriguing that the adversarial perturbations transfer well within the BNNs and can severely degrade the FP32 accuracy. While the perturbations mounted targeted at the FP32 do not perform well against the BNNs. A closer inspection reveals that the perturbations generated by ReActNet and ReCU exhibit strong transferability towards other networks.


\subsection{Resuts on ImageNet}\label{imagenet}
In this section, we evaluate the robustness of BNNs on ImageNet using the same settings as CIFAR-10.
\begin{figure*}[t]
\begin{minipage}{.57\linewidth}
\begin{center}
\includegraphics[width=1.0\linewidth]{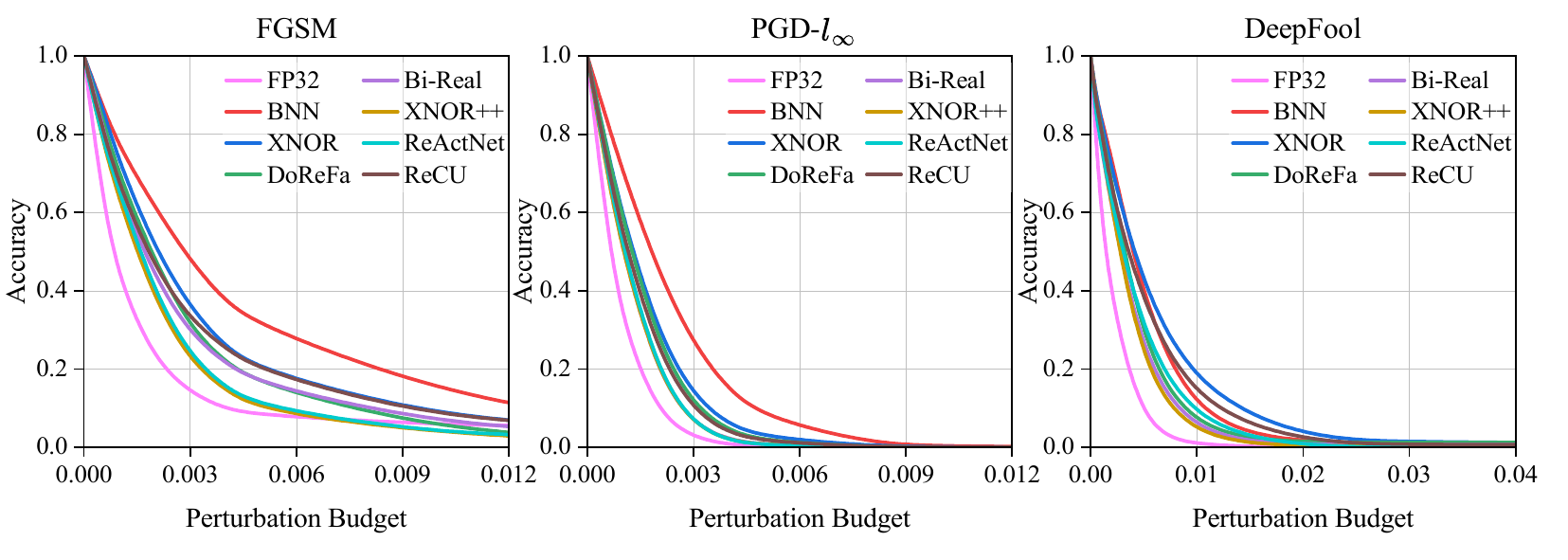}
\end{center}
\vspace{-4ex}
\caption{The \textit{accuracy vs. perturbation budget} curves of the $8$ normally trained models on ImageNet against untargeted white-box attacks under the $\ell_{\infty}$ norm.}
\label{fig:whitelinf-ut-linf-imageNet-acc-pert}
\end{minipage}
\hspace{1ex}
\begin{minipage}{.41\linewidth}
\begin{center}
\includegraphics[width=0.9\linewidth]{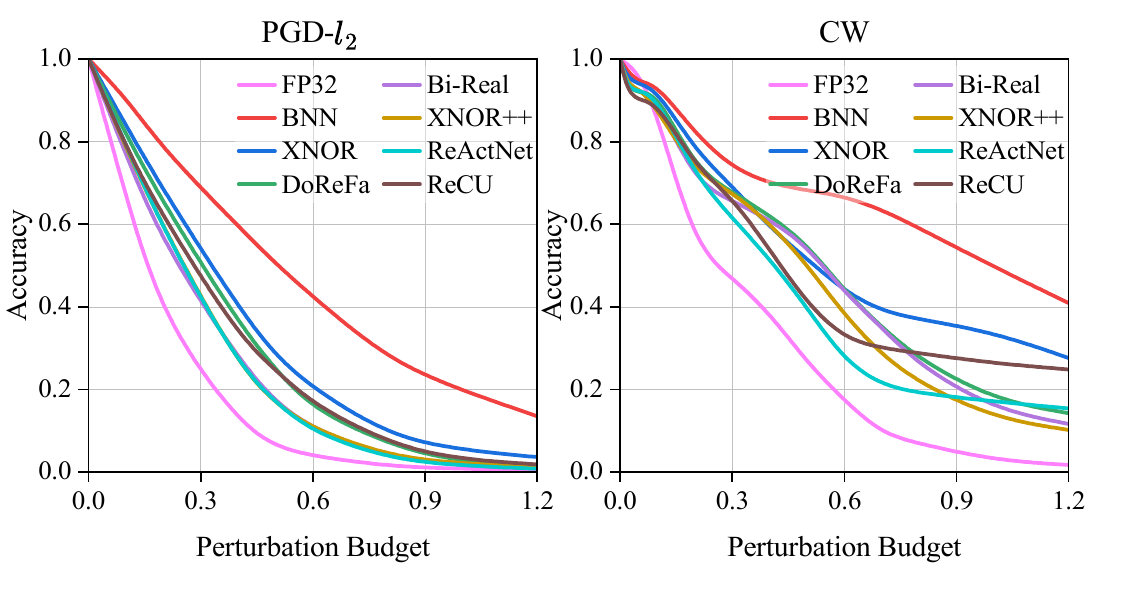}
\end{center}
\vspace{-4ex}
\caption{The \textit{accuracy vs. perturbation budget} curves of the $8$ normally trained models on ImageNet against untargeted white-box attacks under the $\ell_2$ norm.}
\label{fig:whitel2-ut-l2-imageNet-acc-pert}
\end{minipage}
\begin{minipage}{.57\linewidth}
\begin{center}
\includegraphics[width=1.0\linewidth]{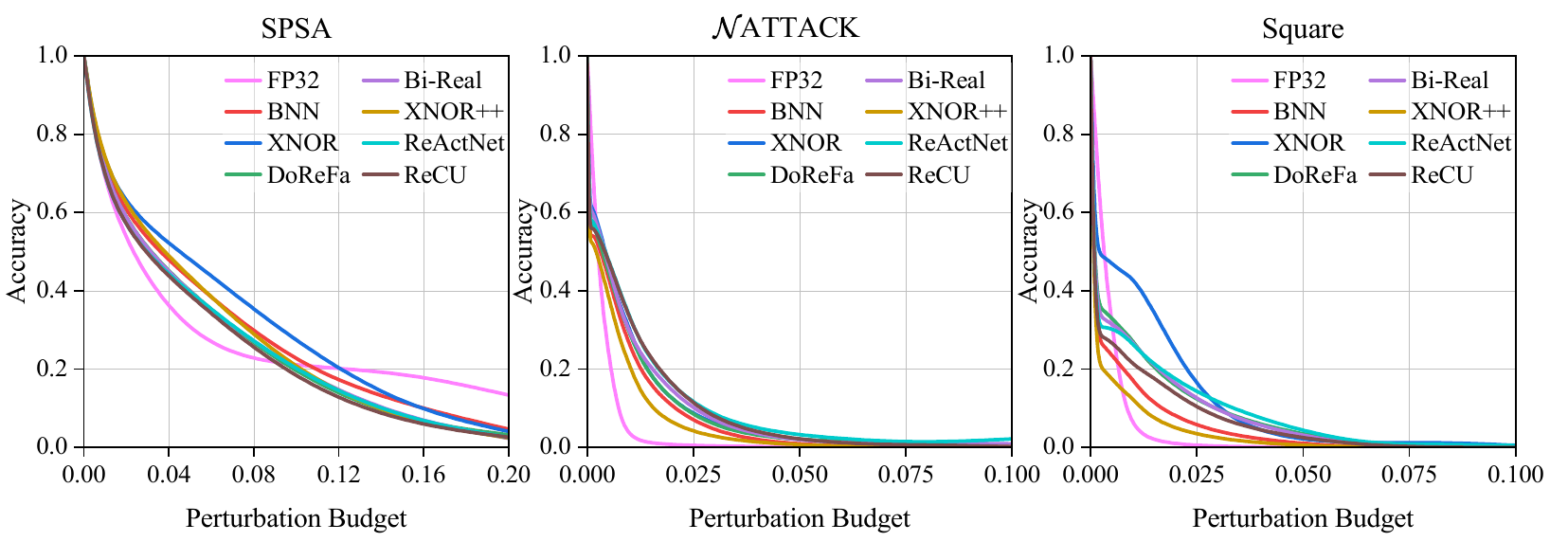}
\end{center}
\vspace{-4ex}
\caption{The \textit{accuracy vs. perturbation budget} curves of the $8$ normally trained models on ImageNet against untargeted score-based attacks under the $\ell_{\infty}$ norm.}
\label{fig:score-ut-linf-imageNet-acc-iter}
\end{minipage}
\hspace{1ex}
\begin{minipage}{.41\linewidth}
\begin{center}
\includegraphics[width=0.9\linewidth]{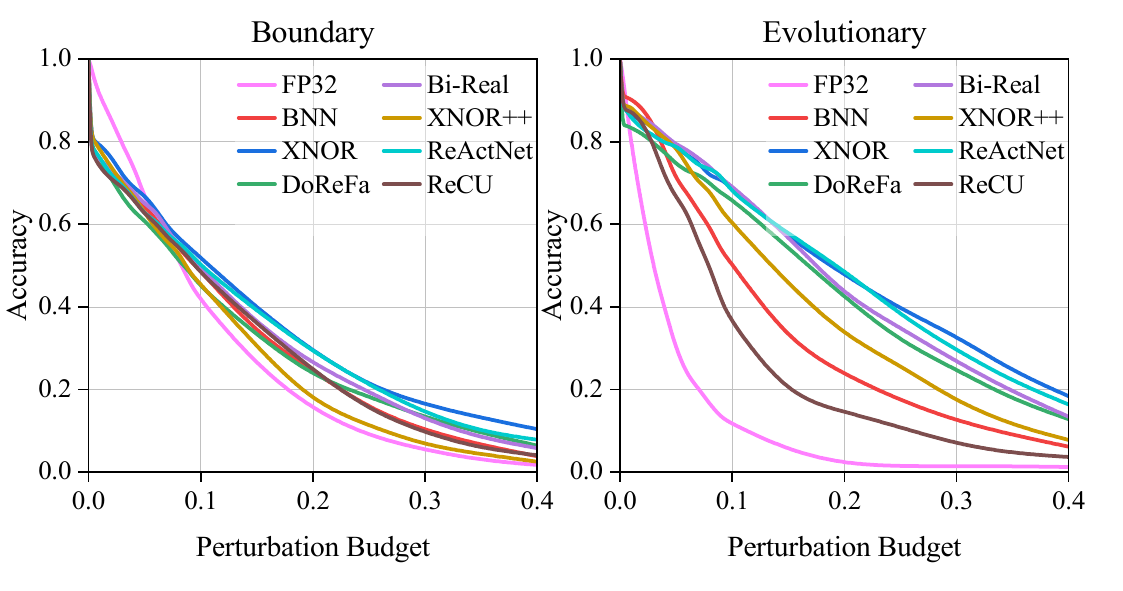}
\end{center}
\vspace{-4ex}
\caption{The \textit{accuracy vs. perturbation budget} curves of the $8$ normally trained models on ImageNet against untargeted decision-based attacks under the $\ell_{\infty}$ norm.}
\label{fig:decision-ut-l2-imageNet-acc-iter}
\end{minipage}
\end{figure*}

\subsubsection{White-box Attacks}
Fig. \ref{fig:whitelinf-ut-linf-imageNet-acc-pert} and Fig. \ref{fig:whitel2-ut-l2-imageNet-acc-pert} illustrate the white-box attack robustness curves of 8 models under the $\ell_{\infty}$ and $\ell_{2}$ norms. When comparing the results on CIFAR-10, we observe that the relative robustness of FP32 and BNNs exhibits the opposite performance. BNNs outperform FP32 in terms of robustness under all perturbations, and BNN \cite{BNN} replaces XNOR++ as the most robust BNNs. While the perturbation is set to the commonly used value of 0.03, the accuracy of BNNs is reduced to 0, making it difficult to make a direct comparison. We set the perturbation for white-box attacks to 0.003 in Tab. \ref{tab:tab2} under $\ell_\infty$ norm. We also observe a significantly higher robustness score for BNN \cite{BNN} compared to other binarization methods and FP32.

\subsubsection{Black-box Attacks}

\textbf{Score-based and Decision-based Attacks}:
The robustness curves based on score-based and decision-based black-box attacks on the ImageNet dataset are presented in Fig. \ref{fig:score-ut-linf-imageNet-acc-iter} and Fig. \ref{fig:decision-ut-l2-imageNet-acc-iter}, respectively. 
 A distinct phenomenon, different from that observed on the CIFAR-10 dataset, can be observed in Fig. \ref{fig:score-ut-linf-imageNet-acc-iter}. The accuracy of the FP32 declines more rapidly to 0 under $\mathcal{N}$attack and Square attacks. This phenomenon is also evident in Fig. \ref{fig:decision-ut-l2-imageNet-acc-iter}. From the point-wise results in Tab. \ref{tab:tab2}, it can be observed that XNOR demonstrates the best overall robustness performance under both score-based and decision-based attacks. 

\begin{figure}[ht]
  \centering
  \vspace{-0.8ex}
  \includegraphics[width=\linewidth]{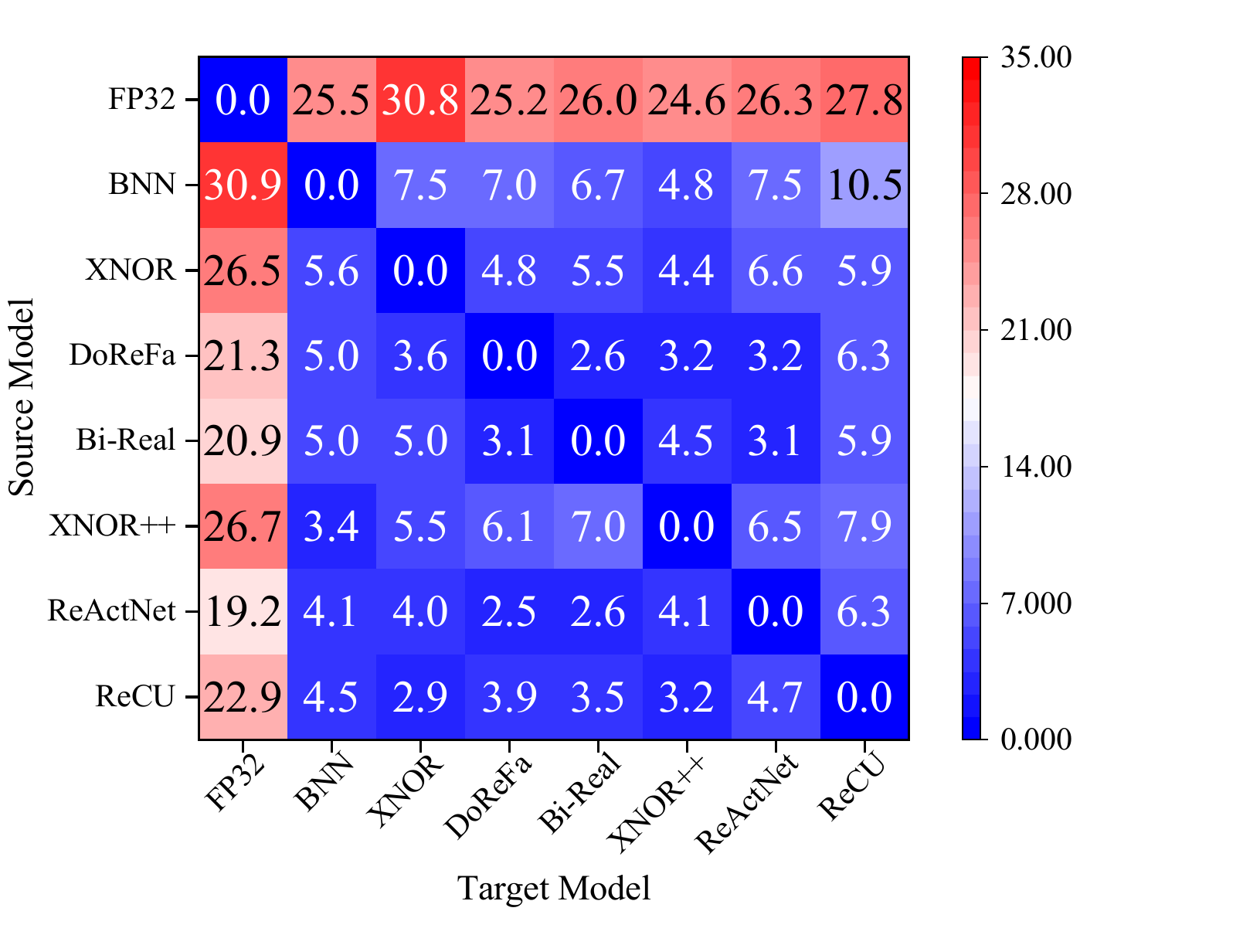}
  \caption{The heatmap of the 8 normally trained models on ImageNet against SI-NI-FGSM attacks under the $\ell_{\infty}$ norm.}
  \label{fig:tranfer_normal2}
\end{figure}

\textbf{Transfer-based Attacks}:
The results of transfer attacks are presented in Fig. \ref{fig:tranfer_normal2}. It can be observed that the transferability between BNNs and FP32 is relatively poor. Conversely, adversarial perturbations transfer well within the BNNs, similar to the results observed on CIFAR-10. This implies that different BNNs tend to learn similar features on the ImageNet, whereas there exist disparities in the features learned by FP32 and BNNs.


\begin{table*}[htbp]
  \centering
    \small
    \begin{tabular}{cccccccccc}
    \toprule
    \multicolumn{2}{c}{Attack} & FP32  & BNN   & XNOR  & DoReFa & Bi-Real & XNOR++ & ReActNet & ReCU \\
    \midrule
    \midrule
    \multirow{6}[2]{*}{White} & FGSM  & 13.96\% & \textbf{47.66\%} & 35.65\% & 30.54\% & 29.05\% & 22.01\% & 23.41\% & 32.71\% \\
          & PGD   & 2.46\% & \textbf{26.33\%} & 13.08\% & 10.59\% & 5.80\% & 5.72\% & 5.82\% & 9.33\% \\
          & DeepFool & 21.87\% & \textbf{60.00\%} & 58.24\% & 49.73\% & 46.03\% & 44.32\% & 49.41\% & 52.13\% \\
          & PGD($\ell_2$) & 6.19\% & \textbf{50.47\%} & 28.15\% & 24.82\% & 16.90\% & 16.60\% & 16.92\% & 24.72\% \\
          & CW    & 26.62\% & \textbf{68.22\%} & 51.65\% & 55.26\% & 54.85\% & 50.28\% & 39.87\% & 40.82\% \\
          & \textbf{Robustness Score} & 14.22\% & \textbf{50.54\%} & 37.35\% & 34.19\% & 30.53\% & 27.79\% & 27.09\% & 31.94\% \\
    \midrule
    \multirow{4}[2]{*}{Score} & SPSA  & 32.23\% & 53.64\% & 57.51\% & 49.91\% & 48.68\% & \textbf{58.47\%} & 48.07\% & 49.91\% \\
          & $\mathcal{N}$ATTACK & 0.29\% & 2.06\% & 3.85\% & 3.03\% & 3.88\% & 1.49\% & 5.36\% & \textbf{4.45\%} \\
          & Square & 0.14\% & 2.80\% & 3.85\% & 7.84\% & 7.41\% & 1.49\% & \textbf{10.39\%} & 5.57\% \\
          & \textbf{Robustness Score} & 10.89\% & 19.50\% & \textbf{21.73\%} & 20.26\% & 19.99\% & 20.48\% & 21.27\% & 19.98\% \\
    \midrule
    \multirow{3}[2]{*}{Decision} & Boundary & \textbf{79.31\%} & 69.78\% & 71.88\% & 66.61\% & 70.55\% & 69.87\% & 69.46\% & 69.02\% \\
          & Evolutionary & 49.46\% & 83.94\% & 83.36\% & 79.61\% & \textbf{84.03\%} & 83.10\% & 81.50\% & 79.11\% \\
          & \textbf{Robustness Score} & 64.38\% & 76.86\% & \textbf{77.62\%} & 73.11\% & 77.29\% & 76.48\% & 75.48\% & 74.06\% \\
    \bottomrule
    \end{tabular}%
  \caption{The point-wise results of 8 normally trained models on ImageNet against white-box attacks under $\ell_\infty$ norm for fixed $\epsilon=0.003$ and $\ell_2$ norm for fixed $\epsilon=0.5$ and black-box attacks under $\ell_\infty$ norm for fixed $\epsilon=0.03$.}
  \label{tab:tab2}%
\end{table*}%

%% file: sec/4_discussion.tex
\section{Discussion}
Based on the extensive experiments conducted on CIFAR-10 and ImageNet datasets, we have observed two intriguing phenomena. \textbf{1) Opposite robustness performance under white-box attacks}. Specifically, on CIFAR-10, BNNs exhibit lower robustness compared to FP32 models against all perturbations. However, on ImageNet, the relationship is entirely reversed, with BNNs demonstrating superior robustness. \textbf{2) Better robustness performance under black-box attacks}. Regardless of the dataset, BNNs exhibit enhanced robustness under score-based and decision-based attacks. In this section, we further explore the adversarial robustness of BNNs based on these notable findings.
\subsection{Exploration under White-Box Attacks}
In this section, we investigate the factors that influence the contrasting robustness performance under white-box attacks. Considering the differences in datasets, we hypothesize that the opposite robustness evaluation results come from \textbf{number of images per category}, \textbf{image resolution}, and \textbf{number of categories}. We analyze each aspect separately to gain insights into this phenomenon.
To explore this issue, we constructed a dataset called ImageNet-10, which has the same number of categories and data volume as CIFAR-10, based on ImageNet. The specific details of the dataset are provided in Appendix D. In the following experiments, We use BNN \cite{BNN} as the binarization method and PGD as the attack method.

\begin{figure}[ht]
  \centering
  \includegraphics[width=\linewidth]{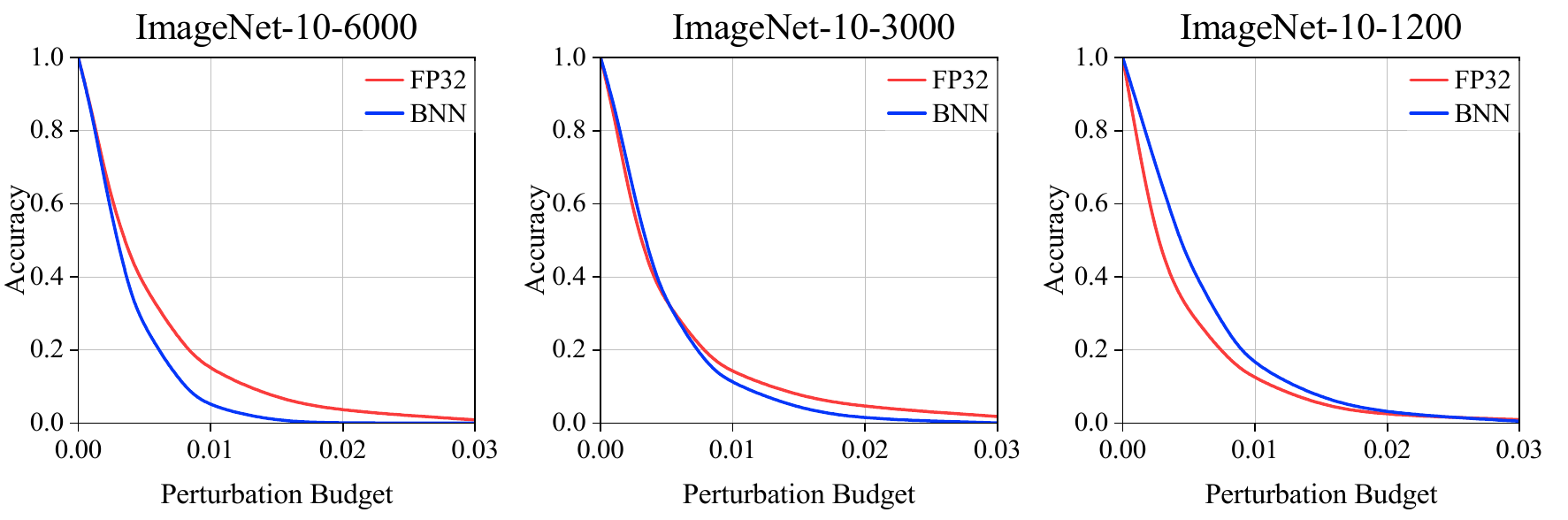}
  \caption{The robustness curves of FP32 and BNN against PGD attacks under $\ell_{\infty}$ norm. ``ImageNet-10-6000" means that the number of images per category is 6,000 on the ImageNet-10 dataset.}
  \label{fig:discussion1}
  
\end{figure}
\textbf{Number of Images per Category}:
We first investigate the impact of the number of images per category on the robustness between BNN and FP32. By controlling the number of images per class in the ImageNet-10 dataset to be 6,000, 3,000, and 1,200, we retrain FP32 and BNN models in the same setting. The robustness curves under PGD attack are presented in Fig. \ref{fig:discussion1}. It is evident that the relative robustness of FP32 and BNN gradually changes with the variation of dataset volume. Specifically, when the number of images per category is 6,000, FP32 exhibits greater robustness than BNN. However, as the number of images per class decreases, the robustness of BNN surpasses that of FP32. Therefore, within the scope of our experiments, we conduct that as the number of images per category decreases from 6,000 (similar to CIFAR-10) to 1,200 (similar to ImageNet), the robustness of BNN gradually strengthens under white-box attacks.

\begin{figure}[ht]
  \centering
  \includegraphics[width=\linewidth]{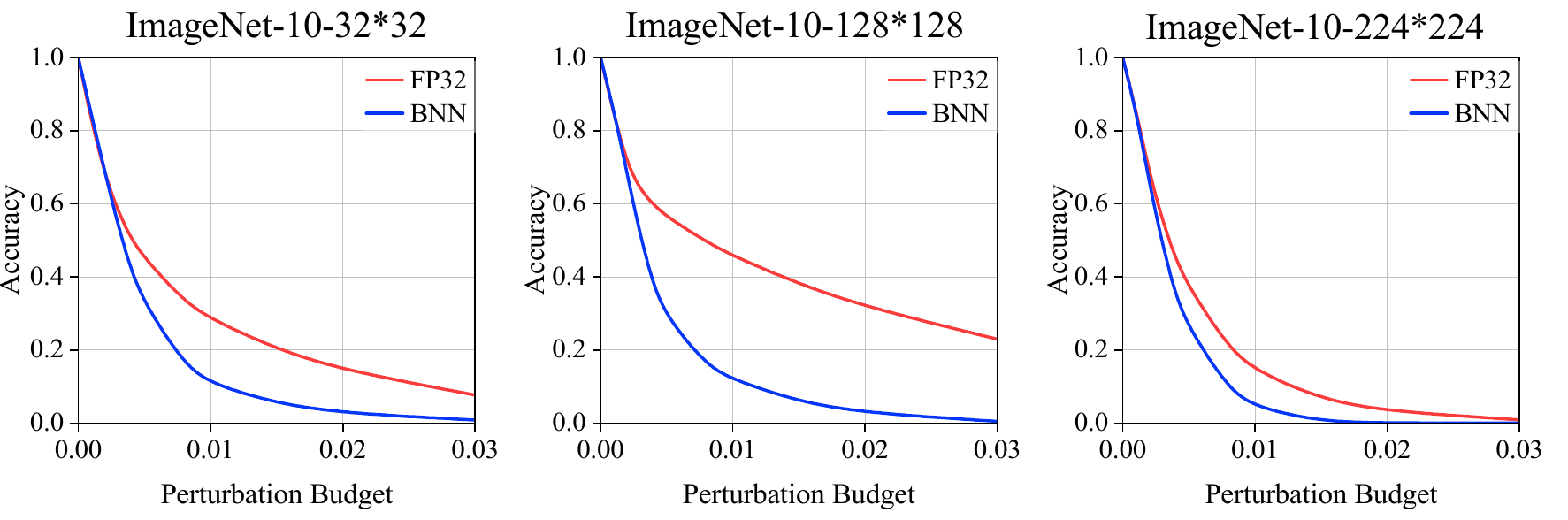}
  \caption{The robustness curves of FP32 and BNN against PGD attacks under $\ell_{\infty}$ norm. ``ImageNet-10-32*32" means that the resolution of the images is 32*32 on the ImageNet-10 dataset.}
  \label{fig:discussion3}
\end{figure}

\textbf{Image Resolution}:
We scale the image resolution of the ImageNet-10 dataset to 224*224, 128*128, and 32*32, respectively. The robustness curves are shown in Fig. \ref{fig:discussion3}. It is evident that FP32 is more robust than BNN at different image resolutions. Particularly, there is a significant gap in accuracy between FP32 and BNN when the image resolution is 128*128. Therefore, we conclude that resolution is not a significant factor in altering the robustness of BNN.

\begin{figure}[ht]
  \centering
  \includegraphics[width=\linewidth]{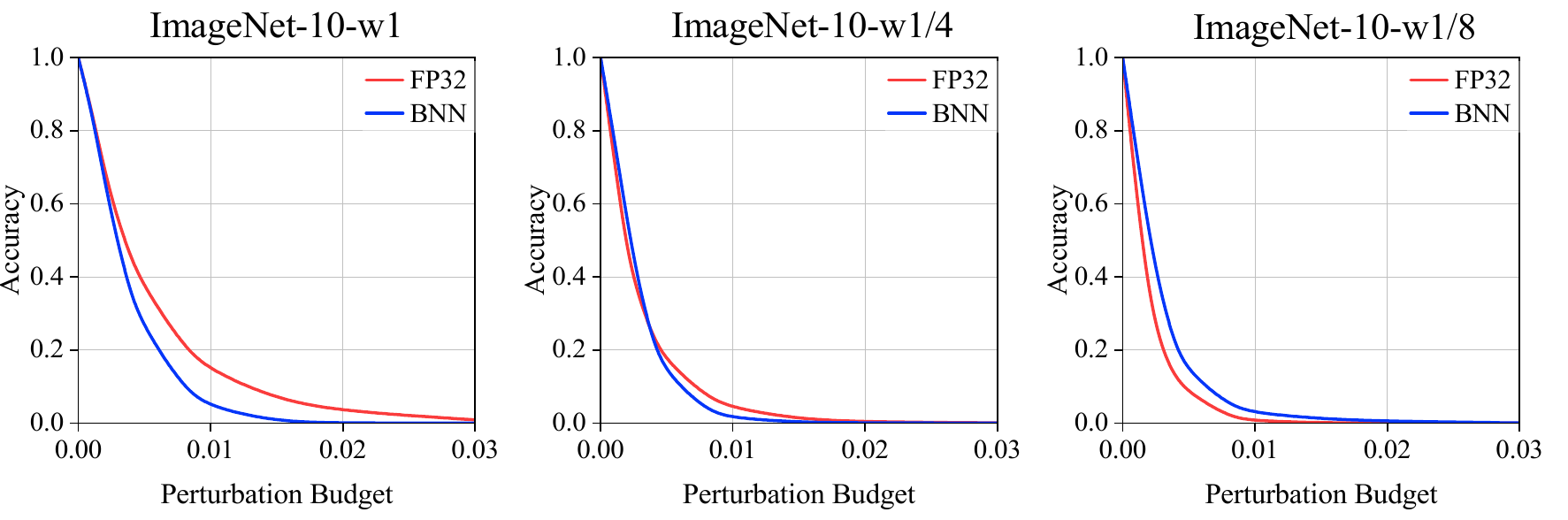}
  \caption{The robustness curves of FP32 and BNN against PGD attacks under $\ell_{\infty}$ norm. ``ImageNet-10-w1/4" means that the model used on the ImageNet-10 dataset has its width reduced to one-fourth of its original width.}
  \label{fig:discussion2}
\end{figure}

\textbf{Number of Categories}:
As the number of categories increases, the capacity of both networks gradually becomes insufficient to produce discriminative features. However, since it is not feasible to construct a large number of categories on ImageNet while ensuring that each category has 6,000 images, we conducted experiments by keeping the dataset categories constant and reducing the capacity of the network. We conduct training and testing on the ImageNet-10 dataset by reducing the width of both networks to 1/4 and 1/8 of their original width. As shown in Fig. \ref{fig:discussion2}, we can observe that when the network width is reduced to 1/4,  the disparity between the robustness curves of the two networks becomes minimal. Furthermore, when the width is reduced to 1/8, the robustness of BNN surpasses that of FP32. This experimental result indirectly validates that during the transition from CIFAR-10 to ImageNet, when both networks are unable to handle more categories effectively, BNN gradually exhibits better robustness.

In conclusion, the results present that the robustness of BNN varies and, within the scope of our experiments, increases with a decrease in the number of images per class. Furthermore, as the model capacity decreases, the relative robustness of BNN also becomes stronger.
\subsection{Exploration under Black-Box Attacks}
In this section, we investigate the factors that contribute to the better robustness of BNN observed under both score-based and decision-based black-box attacks. Unlike white-box attacks, black-box attacks can only obtain limited information through many queries. Inspired by \cite{attack_BNN5}, we utilize Class Activation Maps (CAM) to explore the region of interests (RoIs) for the prediction class using clean images (See Fig. \ref{fig:discussion4}). More CAM samples of other BNNs and the CAM results on CIFAR-10 can be found in Appendix E. CAM can highlight the pixels in an image that have a significant impact on the prediction of the network. In CAM visualization, red indicates a higher contribution to the predicted class, while green represents a lower contribution. From Fig. \ref{fig:discussion4}, we observe that the CAM outputs of BNN exhibit more concentrated RoIs than FP32, indicating that smaller regions have a great impact on the final classification. Under white-box attacks, adversaries can easily identify these regions and introduce perturbations to achieve successful attacks. However, under black-box attacks, adversaries cannot directly access the RoIs. With the same number of queries, the more concentrated RoIs in BNN reduce the chances of finding and perturbing the smaller set of critical pixels by the attack model. Consequently, the performance of small and concentrated RoIs in BNN may contribute to their enhanced robustness against black-box attacks.


\begin{figure}[ht]
  \centering
  \includegraphics[width=\linewidth]{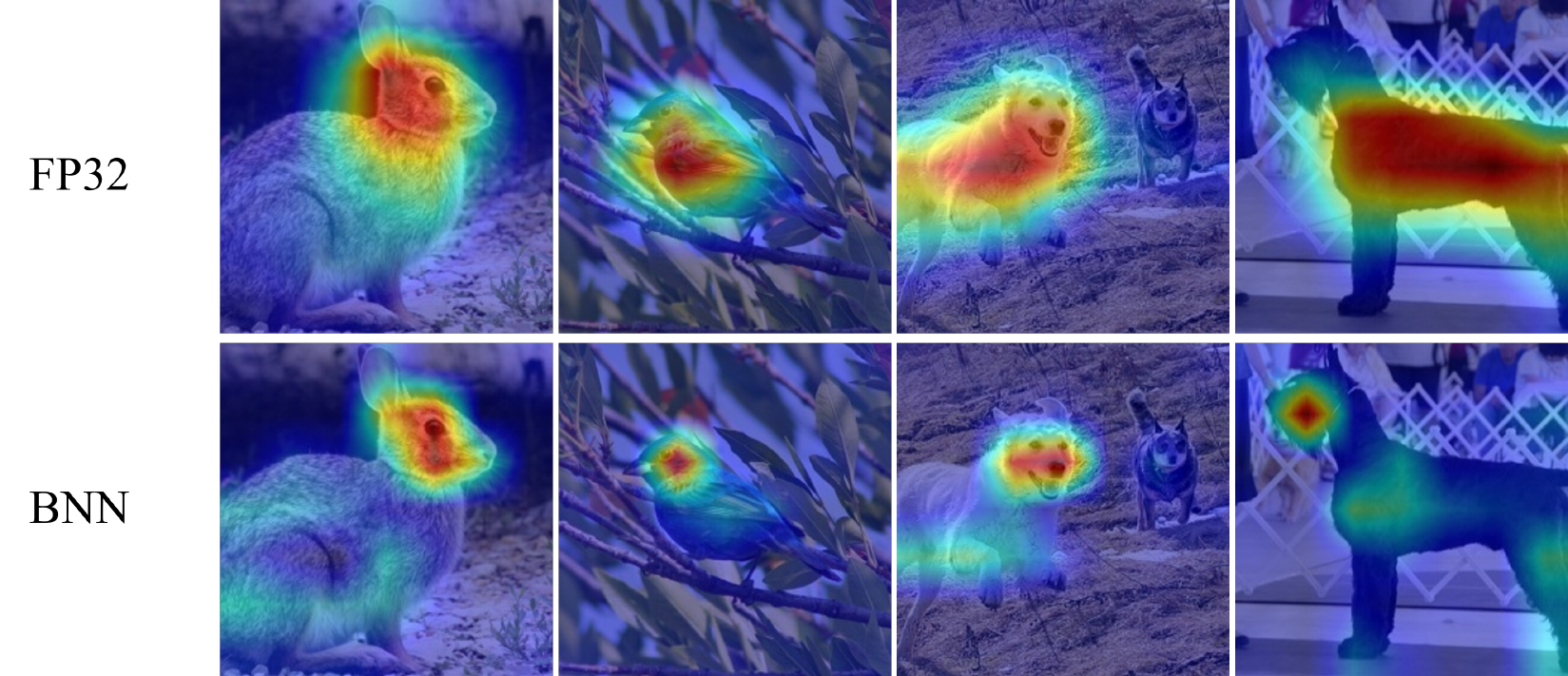}
  \caption{Class activation maps of FP32 and BNN on ImageNet.}
  \label{fig:discussion4}
\end{figure}

\section{Conclusion}

In conclusion, we have presented ARBiBench, a benchmark for evaluating the adversarial robustness of BNNs. We conducted extensive experiments to facilitate a better understanding of the robustness of BNNs. The comprehensive results provide some valuable insights and challenge the previous singular view on the robustness of BNNs. Some of the key observations are as follows: 1) BNNs have greater adversarial robustness than their floating-point counterparts under black-box attacks; 2) BNNs exhibit a completely opposite performance on small-scale and large-scale datasets compared with floating-point counterparts under white-box attacks. We hope our benchmark will inspire future research on enhancing the robustness of BNNs and advancing their application in real-world scenarios.

%% file: sec/6_appendix.tex
\onecolumn

\noindent \begin{center} {\large  \textbf{Supplementary Materials of ARBiBench}} \end{center}
\appendix
\section{Details of BNNs}
In Sec. 3.1, We utilize the following binarized models: XNOR-Net, DoReFa, Bi-Real Net, XNOR-Net++, ReActNet, and ReCU. Here, we provide a detailed description of these models. During the training process of a binarized model, it is common to employ the sign function in the forward pass, while gradient approximations such as STE are utilized during the backward propagation. For each binarized model, we provide a detailed description of the inference computation process for weight and activation during training, as well as the function graph for gradient approximations and the gradient approximations derivatives. The details are presented in Fig. \ref{fig:fig1}.

\section{Attack Results on Defense Ways}
Researchers have proposed numerous adversarial defense strategies, mainly divided into \textbf{adversarial training} \cite{pgd-at, trades}, \textbf{input transformation} \cite{jpeg}, \textbf{randomization} \cite{RP}, \textbf{model ensemble} \cite{pang2019improving}, and \textbf{certified defenses} \cite{Dong1}. \textbf{Adversarial training} attempts to improve the robustness of a neural network by augmenting the training data with adversarial examples. \textbf{Input transformation} modifies the inputs before they are fed into the classifier. \textbf{Randomization} can help alleviate the impact of adversarial attacks by randomizing the classifiers. However, some randomization has been demonstrated by \cite{false_gradient} to introduce false robustness. \textbf{Model ensemble} attempts to construct an ensemble of individual models for adversarial defense. \textbf{Certified defenses} are designed to guarantee formal robustness against adversarial perturbations within defined threat models.

Given the lightweight characteristics of BNNs and the superior effectiveness and robustness of adversarial training compared to alternative defense approaches \cite{renkuisurvery}. Consequently, we utilize two adversarial training methods, \textbf{PGD-AT} \cite{pgd-at} and \textbf{TRADES} \cite{trades}, alongside two input transformation methods,  \textbf{JPEG} \cite{jpeg} and \textbf{Bit-Red}\cite{bit-red}, and a randomization approach, referred to as \textbf{R\&P} \cite{RP}.

As observed in our previous results, the robustness of various BNNs exhibits a relatively similar trend. Therefore, we focus our evaluation on the defense robustness of the FP32 and BNN models. We provide robustness curves in Fig. \ref{fig:fp32_defense} and Fig. \ref{fig:bnn_defense}, as well as point-wise results in Tab. \ref{tab:defense}.

\begin{figure*}[b]  
  \centering  
  \begin{subfigure}{\linewidth}  
    \centering  
    \includegraphics[width=\linewidth]{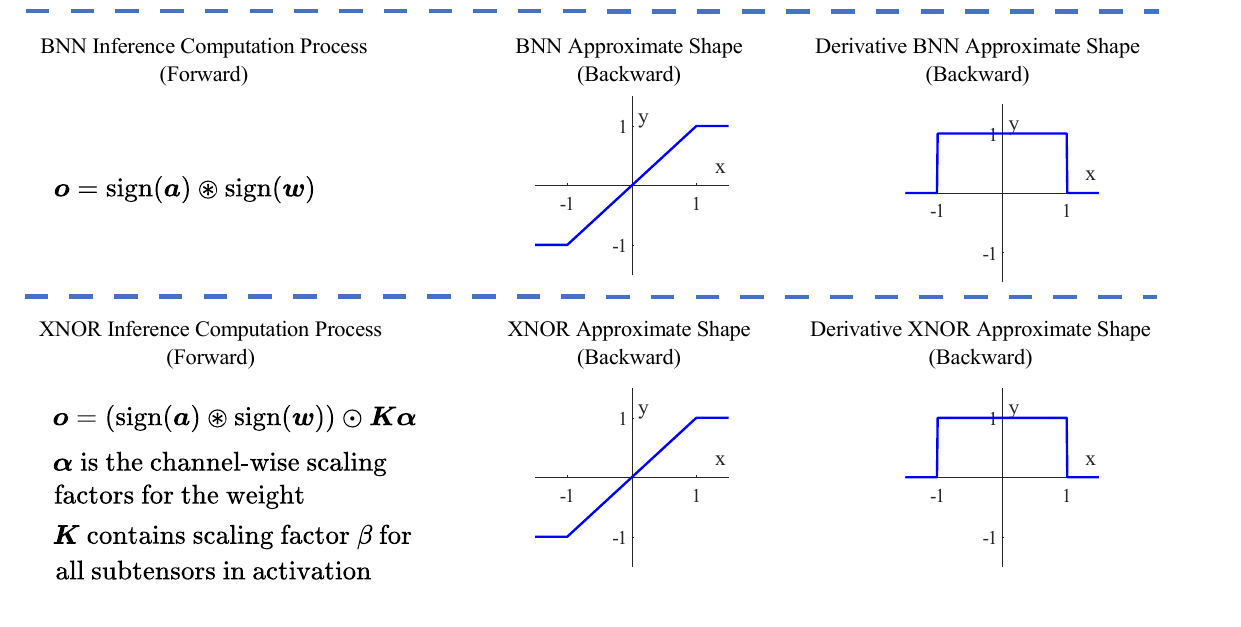}  
    \caption{Forward and backward pass of seven binarized models when training.}  
    \label{fig:bnn1}  
   \vspace{2em}
  \end{subfigure}  
\end{figure*}  

\begin{figure*}[t]  
  \ContinuedFloat  
  \centering  
  \begin{subfigure}{\linewidth}  
    \centering  
    \includegraphics[width=\linewidth]{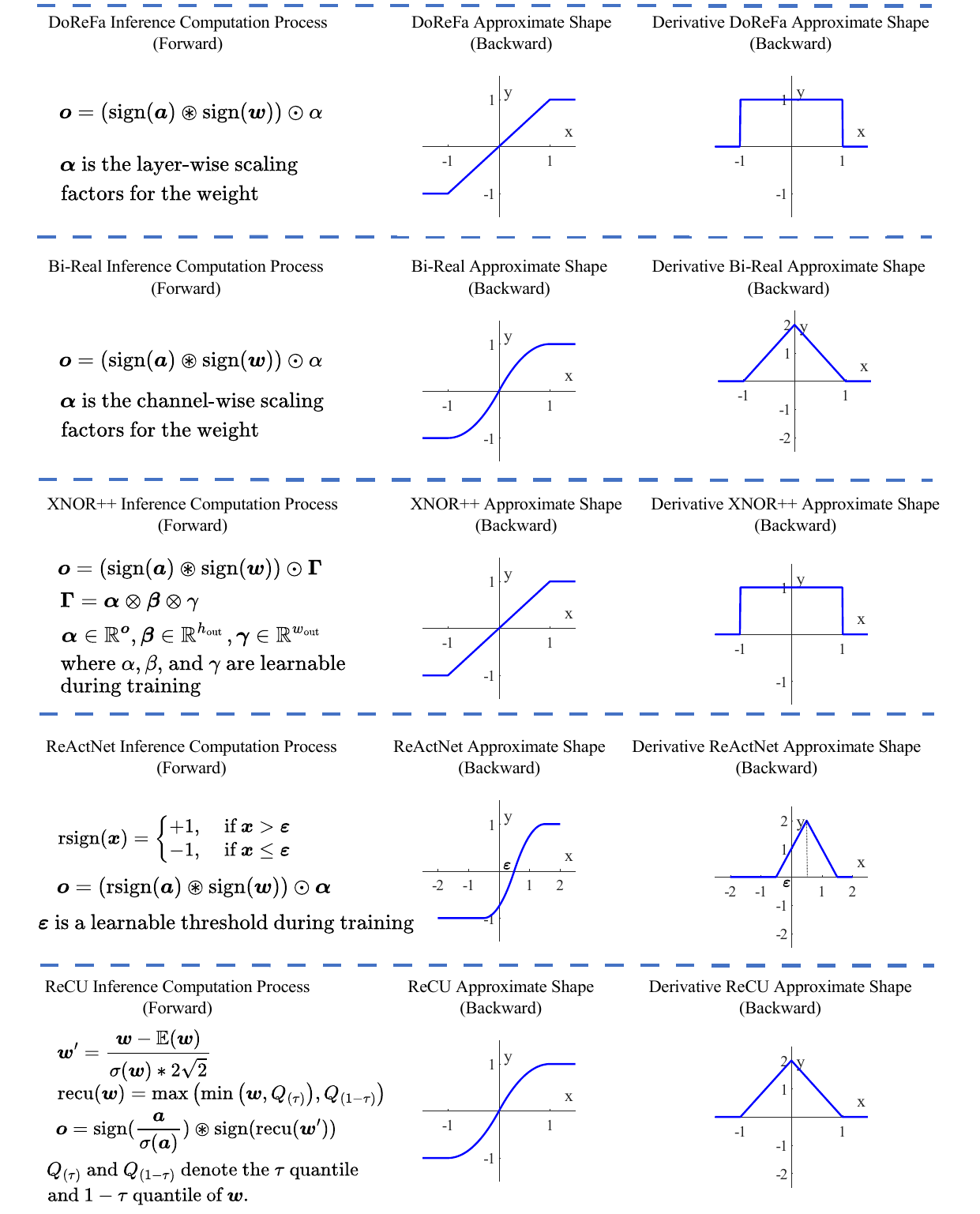}  
    \caption{Forward and backward pass of seven binarized models when training (continued from Figure \ref{fig:bnn1}).}  
    \label{fig:bnn2}  
  \end{subfigure}  
  \caption{Forward and backward pass of seven binarized models when training.} 
  \label{fig:fig1}
\end{figure*}  
\clearpage

\textbf{White-box Attack}: We present the results of FGSM and PGD attacks under $l_\infty$ norm. For both FP32 and BNN, all defense methods improve robustness. From the robustness curves, we observe that adversarial training performs the best among the defense techniques on  BNN, with PGD-AT being more effective. As for the FP32, JPEG defense yields the best results against PGD attacks. We also observe the same phenomenon in the point-wise results.

\begin{table*}[htbp]
  \centering
  
    \begin{tabular}{cccccccc}
    \toprule
    \multicolumn{2}{c}{Defense} & Clean & PGD-AT & TRADES & JPEG  & Bit-Red & R\&P \\
    \midrule
    \midrule
    \multirow{4}[2]{*}{FP32} & FGSM  & 56.68\% & 64.49\% & 68.19\% & \textbf{73.18\%} & 64.70\% & 56.69\% \\
          & PGD   & 12.79\% & 52.80\% & 60.21\% & \textbf{82.12\%} & 45.09\% & 12.36\% \\
          & Square & 18.20\% & 55.29\% & 61.29\% & 82.18\% & \textbf{83.44\%} & 65.73\% \\
          & Boundary & 17.48\% & 22.22\% & 46.81\% & \textbf{83.37\%} & 48.13\% & 75.86\% \\
    \midrule
    \multirow{4}[2]{*}{BNN} & FGSM  & 37.66\% & \textbf{74.61\%} & 70.40\% & 57.33\% & 45.47\% & 39.59\% \\
          & PGD   & 2.83\% & \textbf{68.01\%} & 65.44\% & 48.74\% & 8.16\% & 4.17\% \\
          & Square & 17.48\% & 22.22\% & 46.81\% & \textbf{83.37\%} & 48.13\% & 75.86\% \\
          & Boundary & 63.67\% & 96.24\% & 95.04\% & \textbf{98.48\%} & 91.70\% & 86.33\% \\
    \bottomrule
    \end{tabular}%
    \caption{Point-wise results of 5 defense models on cifar10 against untargeted attacks under $l_\infty$ norm for fixed $\epsilon$ = 0.03.}
  \label{tab:defense}%
\end{table*}%

\textbf{Score-based Black-box Attacks}: We present the results of two models against Square attacks under $l_\infty$ norm. For BNN, the results are entirely different from those observed under white-box attacks. Adversarial training methods exhibit the worst accuracy, with the PGD-AT defense method producing results similar to those without defense. JPEG and R\&P methods prove to be effective. For FP32, all five defense methods deliver commendable results, with JPEG and Bit-Red defenses emerging as the most effective for enhancing robustness.

\textbf{Decision-based Black-box Attacks}: We showcase the results of two models against Boundary attacks under the $l_\infty$ norm. From the robustness curves, a notable observation emerges for both BNN and FP32 models – the three input transformation defense methods exhibit robust performance. Their accuracy remains relatively stable with perturbation increasing. This suggests that even slight perturbations to the input images can render the decision-based attacks ineffective.
\begin{figure*}[ht]
  \centering
  \includegraphics[width=\linewidth]{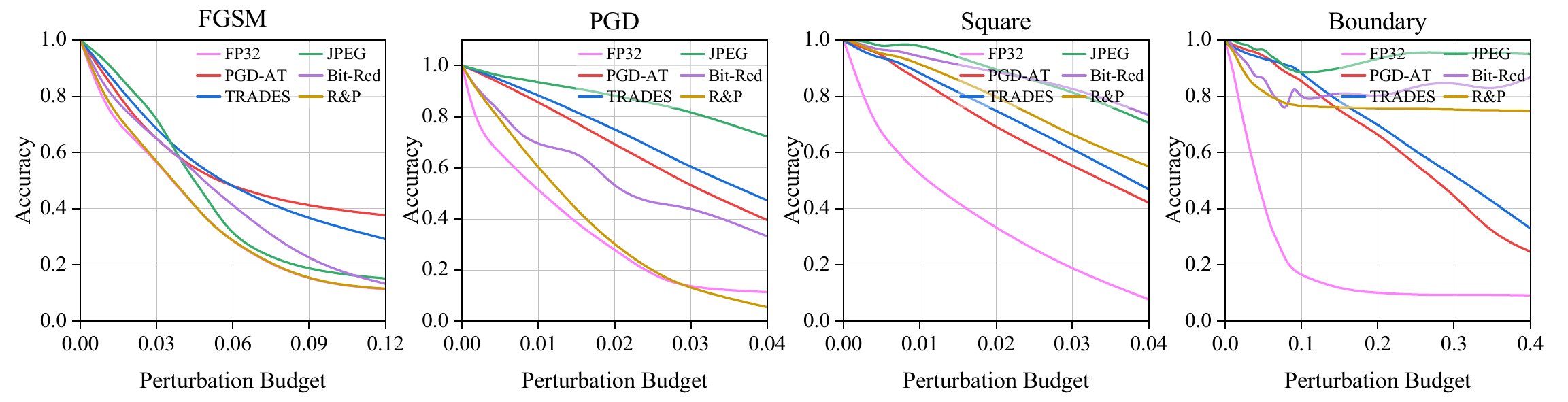}
  \caption{The \textit{accuracy vs. perturbation budget} curves of FP32 with 5 defense ways on CIFAR-10 against untargeted adversarial attacks under the $\ell_{\infty}$ norm.}
  \vspace{-1em}
  \label{fig:fp32_defense}
\end{figure*}

\begin{figure*}[ht]
  \centering
  \includegraphics[width=\linewidth]{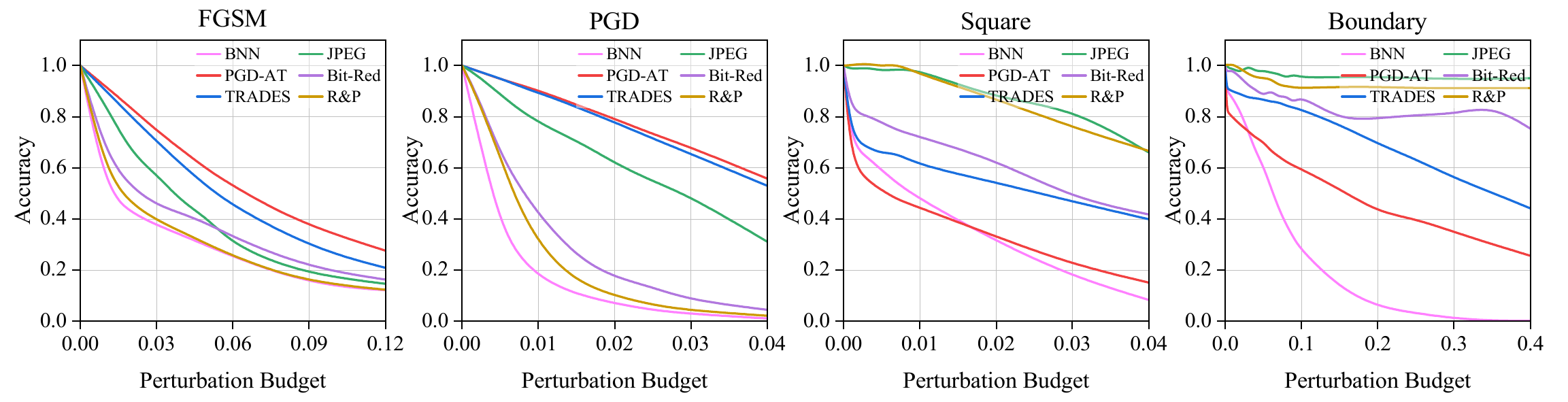}
  \caption{The \textit{accuracy vs. perturbation budget} curves of BNN with 5 defense ways on CIFAR-10 against untargeted adversarial attacks under the $\ell_{\infty}$ norm.}
  \label{fig:bnn_defense}
\end{figure*}

\section{Clean Accuracy of BNNs}
We utilize the same network architecture and binarization method as Bibench. Most of our clean accuracy results were similar to those presented in Bibench, with some of our results on ImageNet surpassing those of Bibench. The results are presented in Tab. \ref{tab:clean acc}.

\begin{table}[htbp]
  \centering
  
    \begin{tabular}{lrrrrrrrr}
    \toprule
    \multicolumn{1}{c}{Datasets} & \multicolumn{1}{c}{FP32} & \multicolumn{1}{c}{BNN} & \multicolumn{1}{c}{XNOR} & \multicolumn{1}{c}{DoReFa} & \multicolumn{1}{c}{Bi-Real} & \multicolumn{1}{c}{XNOR++} & \multicolumn{1}{c}{ReActNet} & \multicolumn{1}{c}{ReCU} \\
    \midrule
    CIFAR-10 & 93.67\% & 90.74\% & 90.31\% & 90.96\% & 91.73\% & 90.92\% & 91.23\% & 91.27\% \\
    ImageNet & 69.29\% & 54.38\% & 56.52\% & 57.19\% & 58.21\% & 54.63\% & 60.14\% & 54.96\% \\
    \bottomrule
    \end{tabular}%
  \caption{Clean accuracy of eight models on CIFAR-10 and ImageNet.}
  \label{tab:clean acc}%
\end{table}%

\section{Details of ImageNet-10}
In Sec. 5.2, we introduce the ImageNet-10 dataset to explore the influencing factors of robustness. Here, we supplement detailed information about the dataset. ImageNet-10 consists of the same number of categories and image quantities as CIFAR-10. The detailed category information is presented in Tab. \ref{tab:Imagnet-10}.

\begin{table}[htbp]
  \centering
  
    \begin{tabular}{cc}
    \toprule
    ImageNet-10 category label & ImageNet category label \\
    \midrule
    \midrule
    fish  &  n01443537 \hspace{0.1cm}   n01484850  \hspace{0.1cm}  n01491361   \hspace{0.1cm} n01494475  \hspace{0.1cm}  n01496331  \\
    bird  &  n01530575 \hspace{0.1cm}   n01531178  \hspace{0.1cm}  n01532829 \hspace{0.1cm}   n01534433  \hspace{0.1cm}  n01537544  \\
    gecko &  n01629819 \hspace{0.1cm}   n01630670 \hspace{0.1cm}   n01631663  \hspace{0.1cm}  n01632458  \hspace{0.1cm}  n01632777  \\
    turtle &  n01664065 \hspace{0.1cm}   n01665541  \hspace{0.1cm}  n01667114  \hspace{0.1cm}  n01667778  \hspace{0.1cm}  n01669191  \\
    snake &  n01728572  \hspace{0.1cm}  n01728920 \hspace{0.1cm}   n01729322  \hspace{0.1cm}  n01729977  \hspace{0.1cm}  n01734418  \\
    spider &  n01773157 \hspace{0.1cm}   n01773549  \hspace{0.1cm}  n01773797  \hspace{0.1cm}  n01774384  \hspace{0.1cm}  n01774750  \\
    dog   &  n02085620 \hspace{0.1cm}   n02087046  \hspace{0.1cm}  n02085936  \hspace{0.1cm}  n02086079  \hspace{0.1cm}  n02086240  \\
    cat   &  n02123045 \hspace{0.1cm}   n02123159 \hspace{0.1cm}   n02123394 \hspace{0.1cm}   n02123597 \hspace{0.1cm}   n02124075  \\
    butterfly &  n02276258 \hspace{0.1cm}   n02277742  \hspace{0.1cm}  n02279972 \hspace{0.1cm}   n02280649  \hspace{0.1cm}  n02281406  \\
    sheep &  n02412080  \hspace{0.1cm}  n02415577 \hspace{0.1cm}   n02417914  \hspace{0.1cm}  n02422106 \hspace{0.1cm}   n02422699  \\
    \bottomrule
    \end{tabular}%
  \caption{The images in each category of ImageNet-10 are selected from five subcategories in ImageNet.}
  \vspace{1em}
  \label{tab:Imagnet-10}%
\end{table}%



\section{CAM Visualization}
\begin{figure*}[ht]
  \centering
  \includegraphics[width=\linewidth]{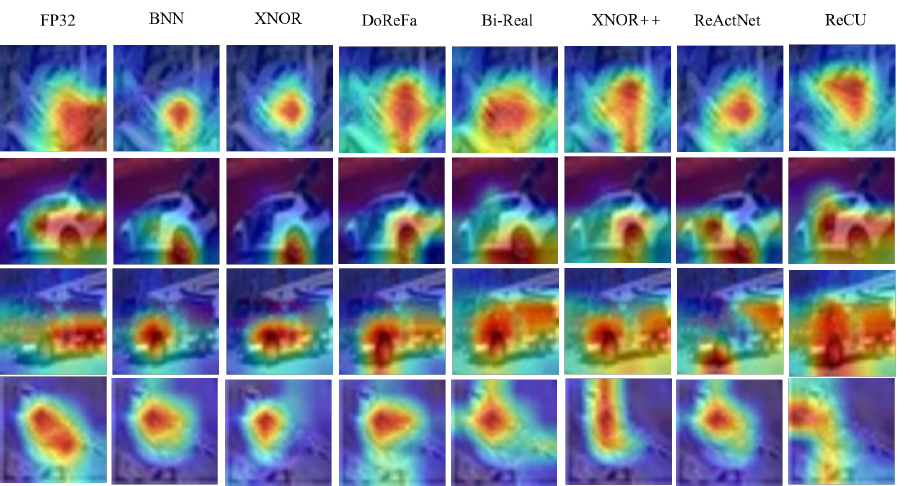}
  \caption{The CAM Visualization on CIFAR-10.}
  \label{fig:cam_cifar}
\end{figure*}

\begin{figure*}[ht]
  \centering
  \includegraphics[width=\linewidth]{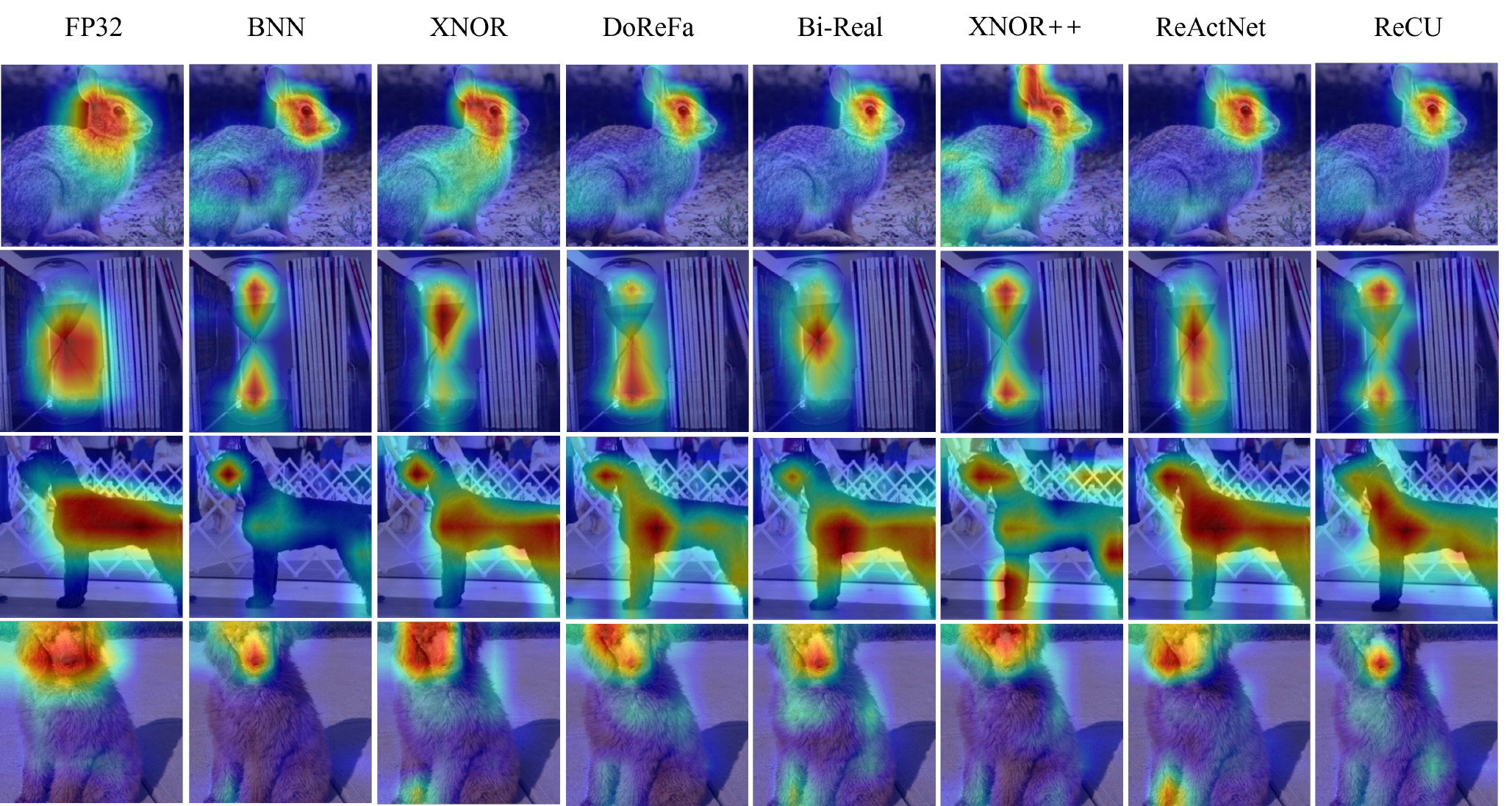}
  \caption{The CAM Visualization on ImageNet.}
  \label{fig:cam_imagenet}
\end{figure*}

In Sec. 5.2, we use Class Activation Mapping (CAM) to attempt to explain the robustness phenomenon of BNNs under black-box attacks. Here, we supplement the CAM visualizations for multiple models. We present the CAM visualization of eight models on CIFAR-10 and ImageNet in Fig. \ref{fig:cam_cifar} and Fig. \ref{fig:cam_imagenet}. 
On ImageNet, we observe that seven BNNs exhibit more concentrated RoIs compared to FP32, and the attention regions of BNNs show some differences compared to FP32. This may be one of the reasons for the relatively poor transferability between FP32 and BNNs. Similar observations were made on CIFAR-10, although the phenomenon was not as pronounced as in the case of ImageNet.

